\documentclass[final]{cvpr}

\usepackage{times}
\usepackage{epsfig}
\usepackage{graphicx}
\usepackage{amsmath}
\usepackage{amssymb}
\usepackage{dsfont}
\usepackage{cuted}
\usepackage{capt-of}
\usepackage{subfigure}
\usepackage{makecell}
\usepackage{epsfig}

\usepackage[pagebackref=true,breaklinks=true,colorlinks,bookmarks=false]{hyperref}

\begin{document}

\title{Make-A-Scene: Scene-Based Text-to-Image Generation with Human Priors}

\author{Oran Gafni\quad Adam Polyak\quad Oron Ashual\quad Shelly Sheynin\quad Devi Parikh\quad Yaniv Taigman\\
Meta AI Research\\
{\tt\small {\{oran,adampolyak,oron,shellysheynin,dparikh,yaniv\}@fb.com}}\vspace{-1cm}}

\maketitle

\begin{strip}\centering
  \centering
  \begin{tabular}{c@{~~~~~~}c}
  \includegraphics[height=6.8cm]{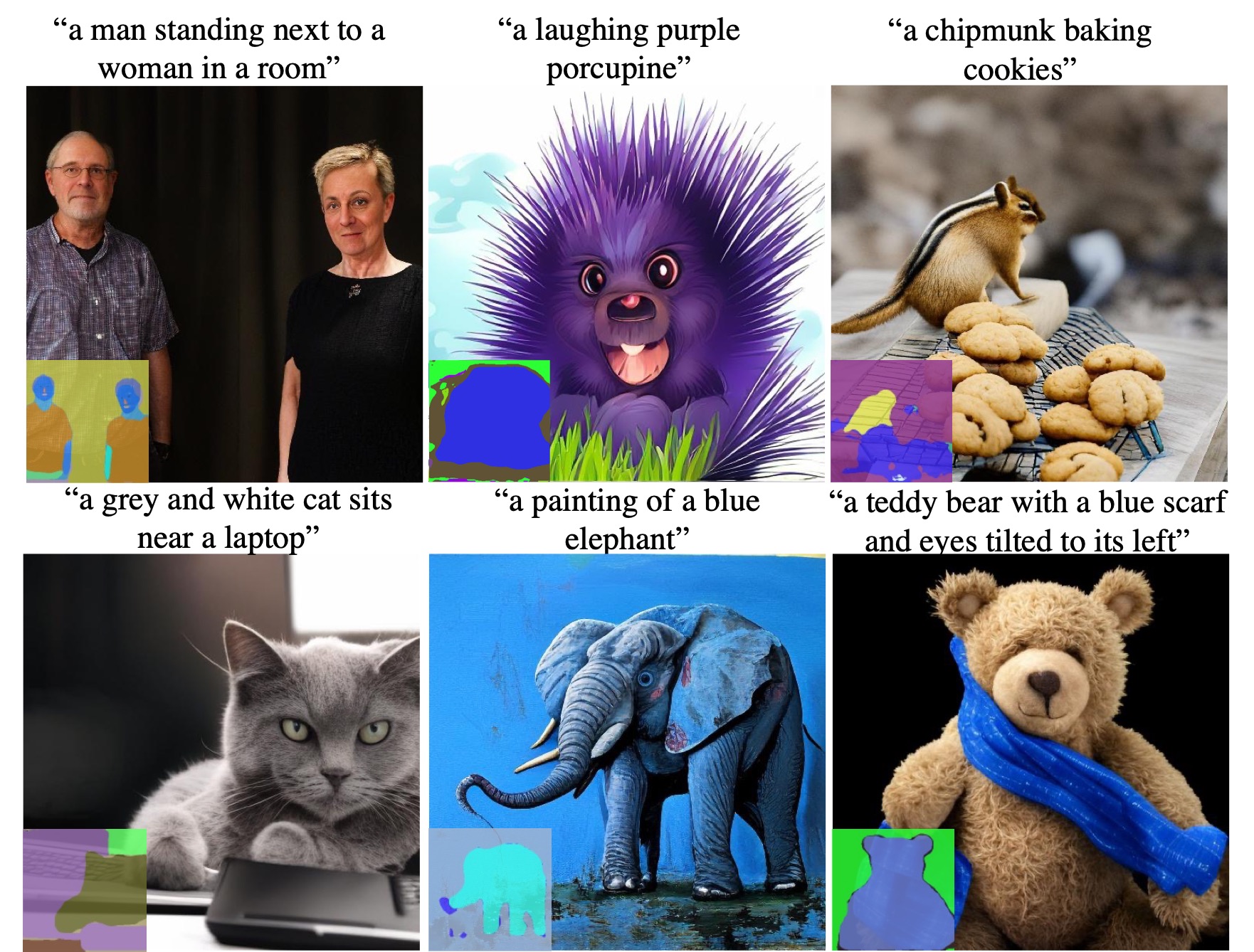} & 
  \includegraphics[height=6.8cm]{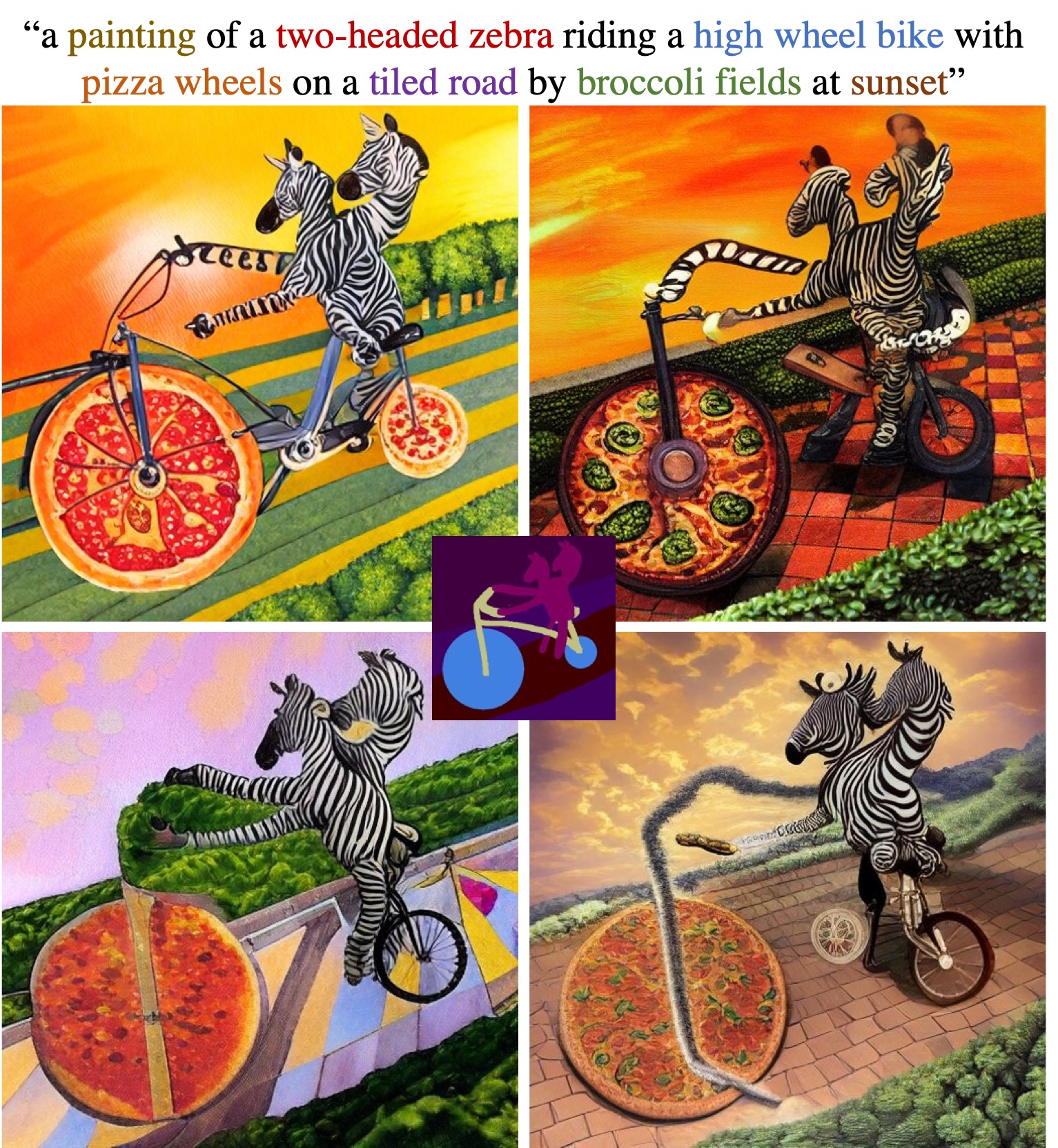} \\
  (a) & (b)
  \end{tabular}
  \captionof{figure}{Make-A-Scene: Samples of generated images from text inputs (a), and a text and scene input (b). Our method is able to both generate the scene (a, bottom left) and image, or generate the image from text and a simple sketch input (b, center).}
  \label{fig:teaser}
\end{strip}

\begin{abstract}\vspace{-0.2cm}
Recent text-to-image generation methods provide a simple yet exciting conversion capability between text and image domains. While these methods have incrementally improved the generated image fidelity and text relevancy, several pivotal gaps remain unanswered, limiting applicability and quality.
We propose a novel text-to-image method that addresses these gaps by (i) enabling a simple control mechanism complementary to text in the form of a scene, (ii) introducing elements that substantially improve the tokenization process by employing domain-specific knowledge over key image regions (faces and salient objects), and (iii) adapting classifier-free guidance for the transformer use case. 
Our model achieves state-of-the-art FID and human evaluation results, unlocking the ability to generate high fidelity images in a resolution of $512\times512$ pixels, significantly improving visual quality. 
Through scene controllability, we introduce several new capabilities: \textbf{(i)} Scene editing, (\textbf{ii}) text editing with anchor scenes, (\textbf{iii}) overcoming out-of-distribution text prompts, and (\textbf{iv}) story illustration generation, as demonstrated in \href{https://youtu.be/QLTyqoJJKTo}{the story we wrote}.
\end{abstract}

\vspace{-0.5cm}
\section{Introduction}
\begin{quote}
    \centering\textit{``A poet would be overcome by sleep and hunger before being able to describe with words what a painter is able to depict in an instant.''}
\end{quote} 

Similar to this quote by Leonardo da Vinci~\cite{janson1991history}, equivalents of the expression ``A picture is worth a thousand words" have been iterated in different languages and eras~\cite{fourcade1968arche,advice1911syracuse,ivan2017fathers}, alluding to the heightened expressiveness of images over text, from the human perspective. There is no surprise then, that the task of text-to-image generation has been gaining increased attention with the recent success of text-to-image modeling via large-scale models and datasets. This new capability of effortlessly bridging between the text and image domains enables new forms of creativity to be accessible to the general public. 

While current methods provide a simple yet exciting conversion between the text and image domains, they still lack several pivotal aspects: 

{\textbf{(i) Controllability}}. The sole input accepted by the majority of models is text, confining any output to be controlled by a text description only. While certain perspectives can be controlled with text, such as style or color, others such as structure, form, or arrangement can only be loosely described at best~\cite{dalleSpotlight}. This lack of control conveys a notion of randomness and weak user-influence on the image content and context~\cite{li2019controllable}. Controlling elements additional to text have been suggested by~\cite{zhang2021m6}, yet their use is confined to restricted datasets such as fashion items or faces. An earlier work by~\cite{hong2018inferring} suggests coarse control in the form of bounding boxes resulting in low resolution images.

{\textbf{(ii) Human perception}}. While images are generated to match human perception and attention, the generation process does not include any relevant prior knowledge, resulting in little correlation between generation and human attention. A clear example of this gap can be observed in person and face generation, where a dissonance is present between the importance of face pixels from the human perspective and the loss applied over the whole image~\cite{judd2012benchmark,yun2013studying}. This gap is relevant to animals and salient objects as well. 

{\textbf{(iii) Quality and resolution}}. Although quality has gradually improved between consecutive methods, the previous state-of-the-art methods are still limited to an output image resolution of $256\times256$ pixels~\cite{ramesh2021zero,nichol2021glide}. Alternative approaches propose a super-resolution network which results in less favorable visual and quantitative results~\cite{ding2021cogview}. Quality and resolution are strongly linked, as scaling up to a resolution of $512\times512$ requires a substantially higher quality with fewer artifacts than $256\times256$.

In this work, we introduce a novel method that successfully tackles these pivotal gaps, while attaining state-of-the-art results in the task of text-to-image generation. Our method provides a new type of control complementary to text, enabling new-generation capabilities while improving structural consistency and quality. Furthermore, we propose explicit losses correlated with human preferences, significantly improving image quality, breaking the common resolution barrier, and thus producing results in a resolution of $512\times512$ pixels.

Our method is comprised of an autoregressive transformer, where in addition to the conventional use of text and image tokens, we introduce implicit conditioning over optionally controlled scene tokens, derived from segmentation maps. During inference, the segmentation tokens are either generated independently by the transformer or extracted from an input image, providing freedom to impel additional constraints over the generated image.
Contrary to the common use of segmentation for explicit conditioning as employed in many GAN-based methods~\cite{isola2017image,wang2018high,park2019semantic}, our segmentation tokens provide implicit conditioning in the sense that the generated image and image tokens are not constrained to use the segmentation information, as there is no loss tying them together. In practice, this contributes to the variety of samples generated by the model, producing diverse results constrained to the input segmentations.

We demonstrate the new capabilities this method provides in addition to controllability, such as (i) complex scene generation (Fig.~\ref{fig:teaser}), (ii) out-of-distribution generation (Fig.~\ref{fig:attending}), (iii) scene editing (Fig.~\ref{fig:edit_scene}), and (iv) text editing with anchored scenes (Fig.~\ref{fig:edit_text}). We additionally provide an example of harnessing controllability to assist with the creative process of storytelling in \href{https://youtu.be/QLTyqoJJKTo}{this video}.

While most approaches rely on losses agnostic to human perception, this approach differs in that respect. We use two modified Vector-Quantized Variational Autoencoders (VQ-VAE) to encode and decode the image and scene tokens with explicit losses targeted at specific image regions correlated with human perception and attention, such as faces and salient objects.
The losses contribute to the generation process by emphasizing the specific regions of interest and integrating domain-specific perceptual knowledge in the form of network feature-matching. 

While some methods rely on image re-ranking for post-generation image filtering (utilizing CLIP~\cite{radford2021learning} for instance), we extend the use of classifier-free guidance suggested for diffusion models~\cite{sohl2015deep,ho2020denoising} by~\cite{ho2021classifier,nichol2021glide} to transformers, eliminating the need for post-generation filtering, thus producing faster and higher quality generation results, better adhering to input text prompts.

An extensive set of experiments is provided to establish the visual and numerical validity of our contributions.

\section{Related Work}
\subsection{Image generation} Recent advancements in deep generative models have enabled algorithms to generate high-quality and natural-looking images. Generative Adversarial Networks~(GANs)~\cite{goodfellow2014generative} facilitate the generation of high fidelity images~\cite{karras2021alias,brock2018large,karras2020analyzing,tseng2021regularizing} in multiple domains by simultaneously training a generator network $G$ and a discriminator network $D$, where $G$ is trained to fool $D$, while $D$ is trained to judge if a given image is real or fake. Concurrently to GANs, Variational Autoencoders~(VAEs)~\cite{kingma2013auto,vahdat2020nvae} have introduced a likelihood-based approach to image generation. Other likelihood-based models include autoregressive models~\cite{van2016conditional,parmar2018image,esser2021taming, pmlr-v119-chen20s} and diffusion models~\cite{dhariwal2021diffusion,ho2022cascaded,ho2020denoising}. While the former model image pixels as a sequence with autoregressive dependency between each pixel, the latter synthesizes images via a gradual denoising process. Specifically, sampling starts with a noisy image which is iteratively denoised until all denoising steps are performed. Applying both methods directly to the image pixel-space can be challenging. Consequently, recent approaches either compress the image to a discrete representation~\cite{esser2021taming,van2017neural} via Vector Quantized~(VQ) VAEs~\cite{van2017neural}, or down-sample the image resolution~\cite{dhariwal2021diffusion,ho2022cascaded}. Our method is based on autoregressive modeling of discrete image representation.

\begin{figure*}
    \centering
    \setlength{\tabcolsep}{2.0pt}
    \begin{tabular}{cc}
        \rotatebox{90}{\phantom{AA} XMC-GAN~\cite{zhang2021cross}} &
        \includegraphics[width=0.98\textwidth]{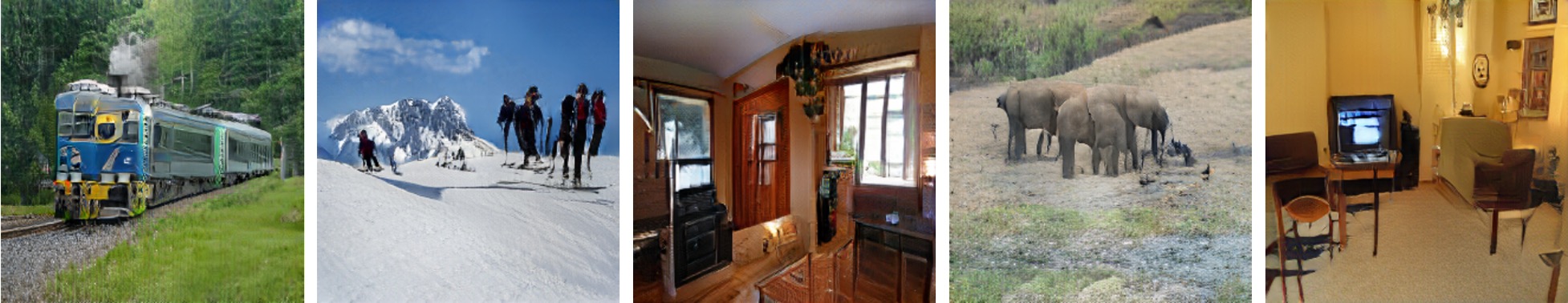} \\
        
        \rotatebox{90}{\phantom{AA} DALL-E~\cite{ramesh2021zero}} &
        \includegraphics[width=0.98\textwidth]{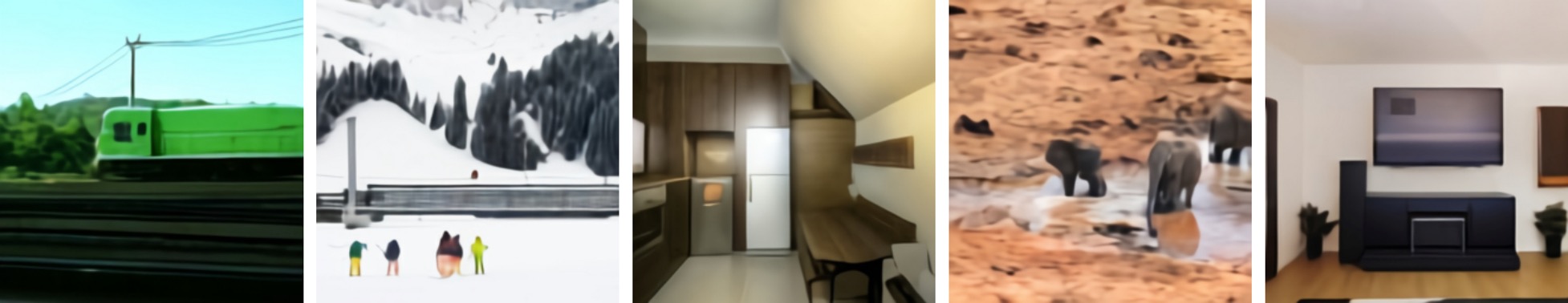} \\

        \rotatebox{90}{\phantom{AA} CogView~\cite{ding2021cogview}} &
        \includegraphics[width=0.98\textwidth]{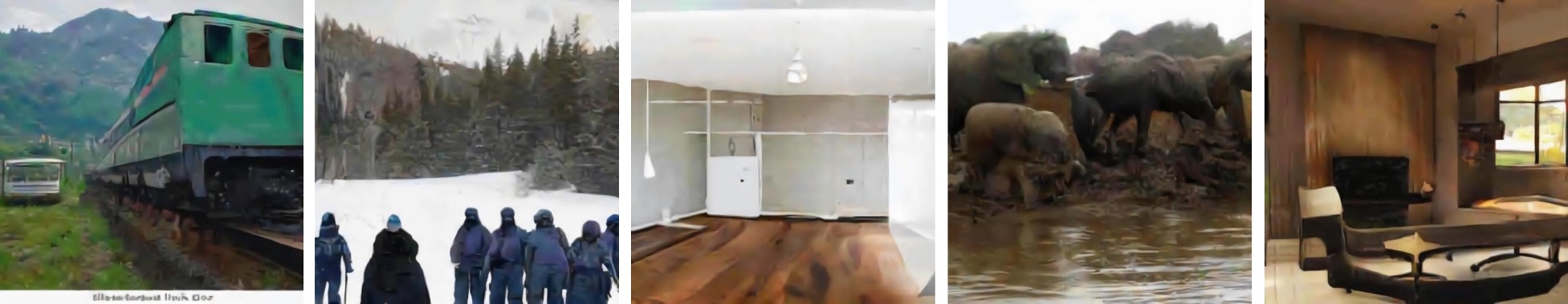} \\
        
        \rotatebox{90}{\phantom{AA} GLIDE~\cite{nichol2021glide}} &
        \includegraphics[width=0.98\textwidth]{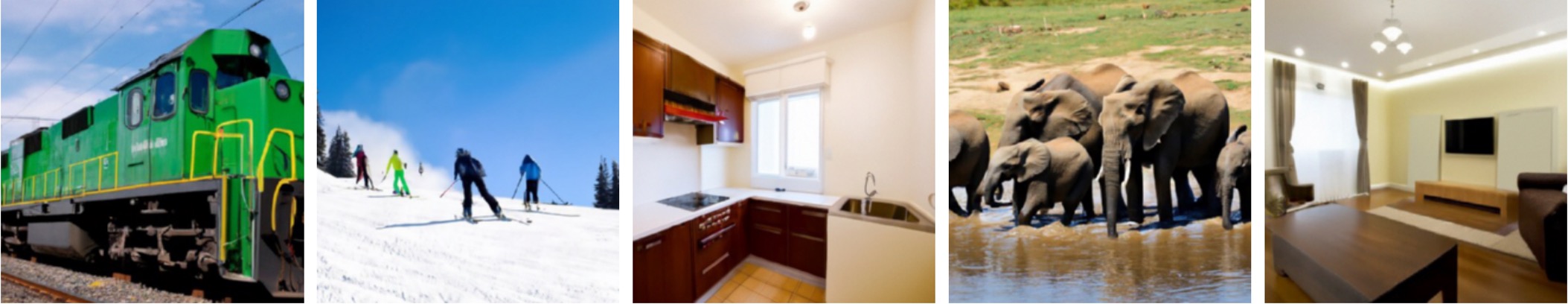} \\

        \rotatebox{90}{\phantom{AAAAA} \textbf{Ours}} &
        \includegraphics[width=0.98\textwidth]{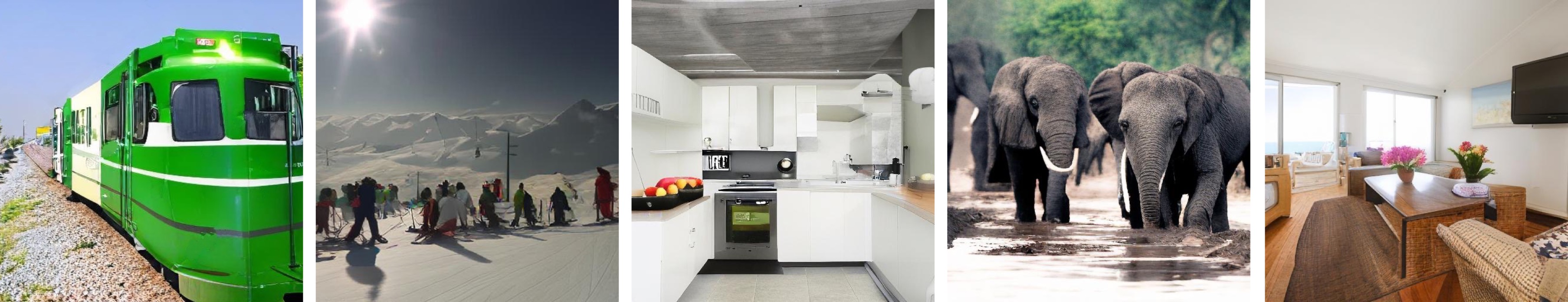} \\

    \end{tabular}
    \begin{tabular}{c@{\hskip 1cm}c@{\hskip 1cm}c@{\hskip 1cm}c@{\hskip 1cm}c@{\hskip 1cm}c@{\hskip 1cm}c}
        & \makecell{``a green train \\ is coming down \\ the tracks''}
        & \makecell{``a group of skiers \\ are preparing to ski \\ down a mountain''}
        & \makecell{``a small kitchen \\ with a \\ low ceiling''}
        & \makecell{``a group of \\ elephants walking \\ in muddy water''}
        & \makecell{``a living area \\ with a television \\ and a table''}
    \end{tabular}
    \vspace{0.3cm}
    \caption{Qualitative comparison with previous work. The text and generated images for~\cite{zhang2021cross,ramesh2021zero,nichol2021glide} were taken from~\cite{nichol2021glide}. For CogView~\cite{ding2021cogview} we use the released $512\times512$ model weights, applying self-reranking of $60$ for post-generation selection.}
    \label{fig:prev_work}
\end{figure*}

\subsection{Image tokenization} Image generation models based on discrete representation~\cite{van2017neural,ramesh2021zero,razavi2019generating,ding2021cogview,esser2021taming} follow a two-stage training scheme. First, an image tokenizer is trained to extract a discrete image representation. In the second stage, a generative model generates the image in the discrete latent space. Inspired by Vector Quantization~(VQ) techniques, VQ-VAE~\cite{van2017neural} learns to extract a discrete latent representation by performing online clustering. 
VQ-VAE-2~\cite{razavi2019generating} presented a hierarchical architecture composed of VQ-VAE models operating at multiple scales, enabling faster generation compared with pixel space generation. The DALL-E~\cite{ramesh2021zero} text-to-image model used dVAE, which uses gumbel-softmax~\cite{jang2016categorical,maddison2016concrete}, relaxing the VQ-VAE's online clustering. Recently, VQGAN~\cite{esser2021taming} added adversarial and perceptual losses~\cite{zhang2018unreasonable} on top of the VQ-VAE reconstruction task, producing reconstructed images with higher quality. In our work, we modify the VQGAN framework by adding perceptual losses to specific image regions, such as faces and salient objects, which further improve the fidelity of the generated images.

\begin{figure*}[t!]
    \centering
    \setlength{\tabcolsep}{2.0pt}
    \begin{tabular}{cc}
        \rotatebox{90}{\phantom{AA} GLIDE~\cite{nichol2021glide}} &
        \includegraphics[width=0.9\textwidth]{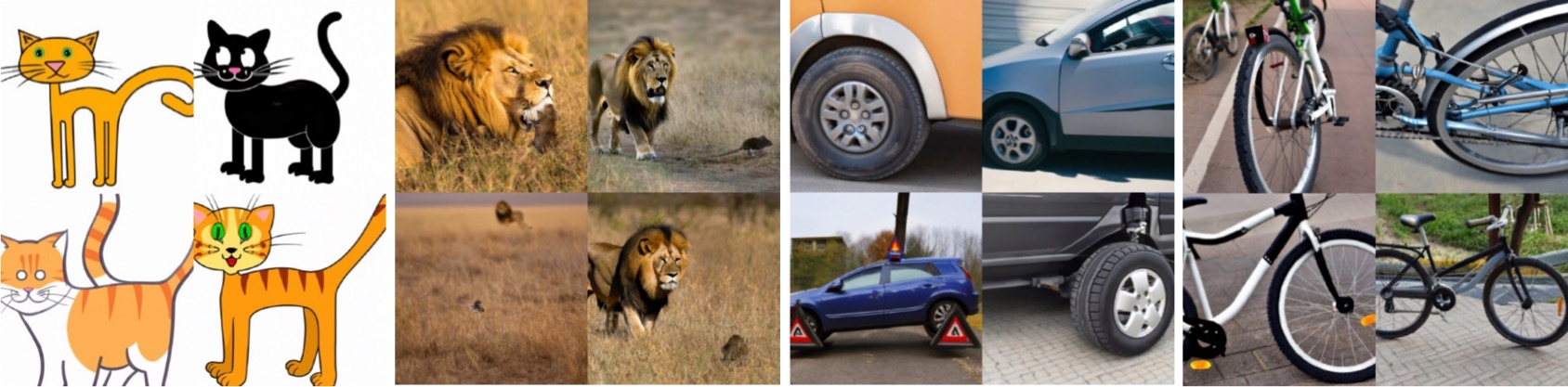} \\

        \rotatebox{90}{\phantom{AAAAAAAAAA} Ours} &
        \includegraphics[width=0.9\textwidth]{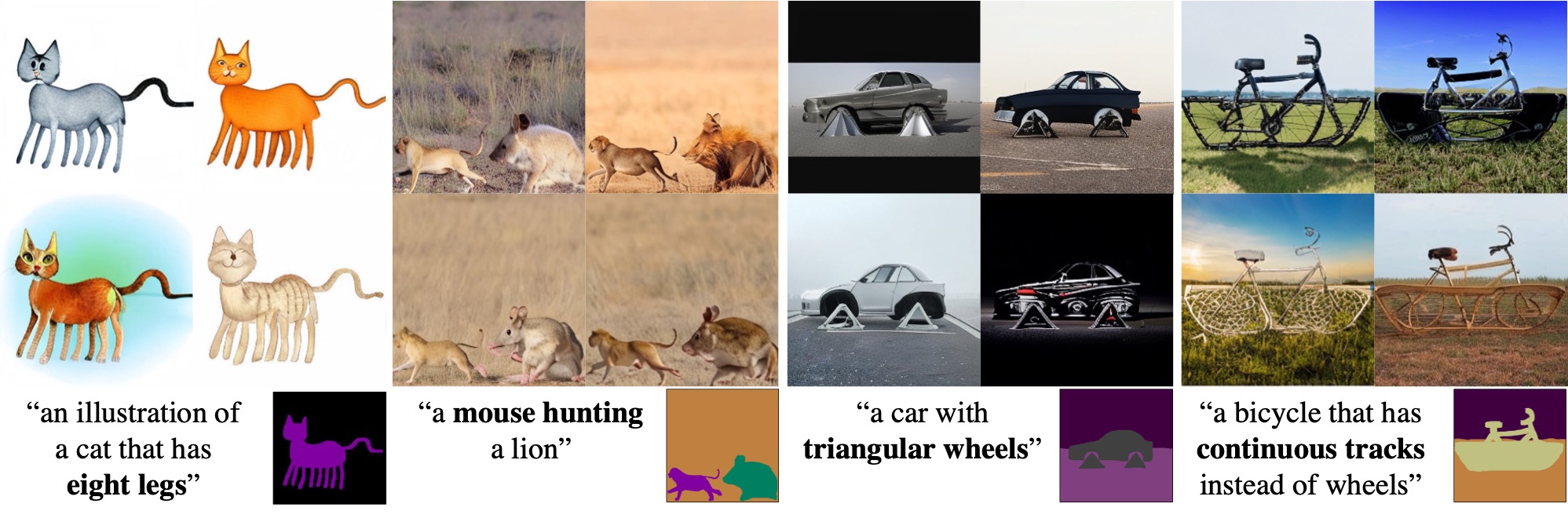} \\

    \end{tabular}
    \caption{Overcoming out-of-distribution text prompts with scene control. By introducing simple scene sketches (bottom right) as additional inputs, our method is able to overcome unusual objects and scenarios presented as failure cases in previous methods.}
    \label{fig:attending}
\end{figure*}

\begin{figure*}[t!]
    \centering
    \setlength{\tabcolsep}{4.0pt}
    \begin{tabular}{ccc@{~~~}c@{}c@{~~~}}
        \includegraphics[height=6.3cm]{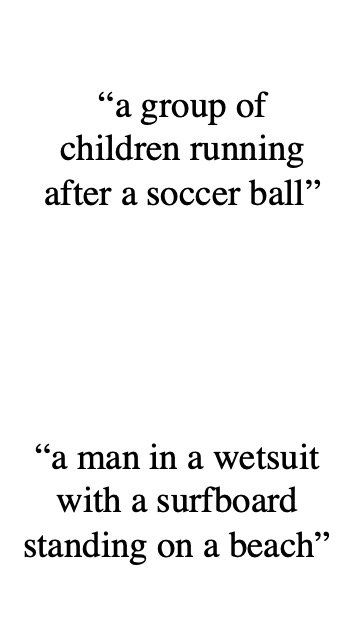} &
        \includegraphics[height=6.6cm]{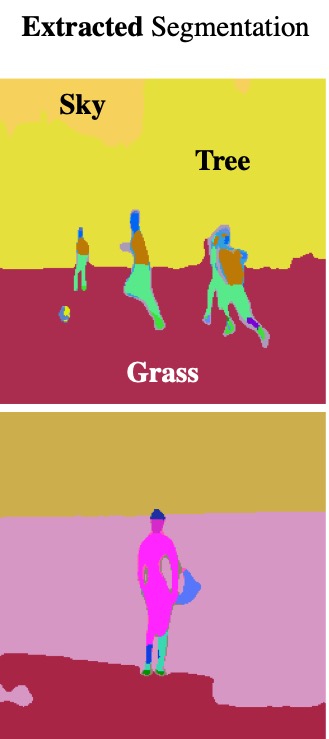} & 
        \includegraphics[height=6.6cm]{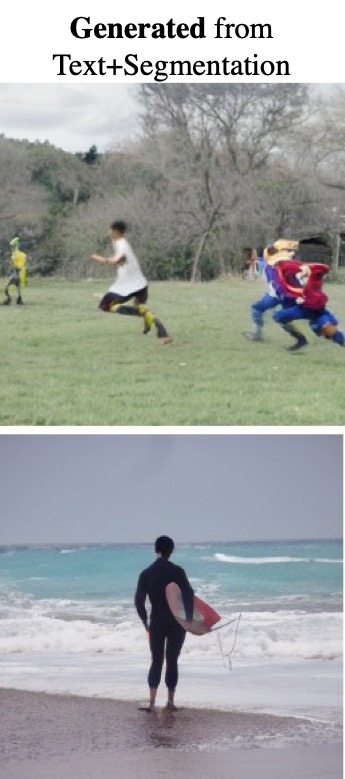} & 
        \includegraphics[height=6.6cm]{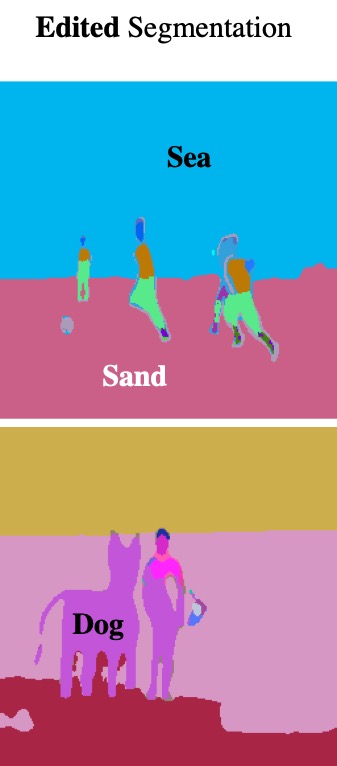} & 
        \includegraphics[height=6.6cm]{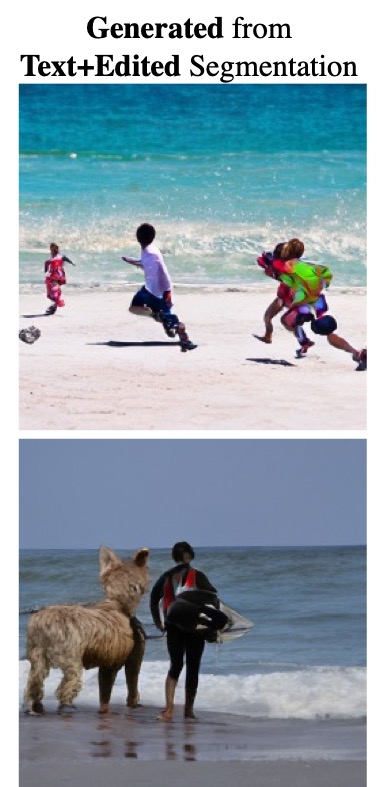} \\
        (a) & (b) & (c) & (d) & (e) 
    \end{tabular}
    \caption{Generating images through edited scenes. For an input text (a) and the segmentations extracted from an input image (b), we can re-generate the image (c) or edit the segmentations (d) by replacing classes (top) or adding classes (bottom), generating images with new context or content (e).}
    \label{fig:edit_scene}
    \vskip -0.1in
\end{figure*}

\begin{figure*}[t!]
    \centering
    \setlength{\tabcolsep}{3.0pt}
    \begin{tabular}{cc@{~~}cc}
        \includegraphics[height=7.2cm]{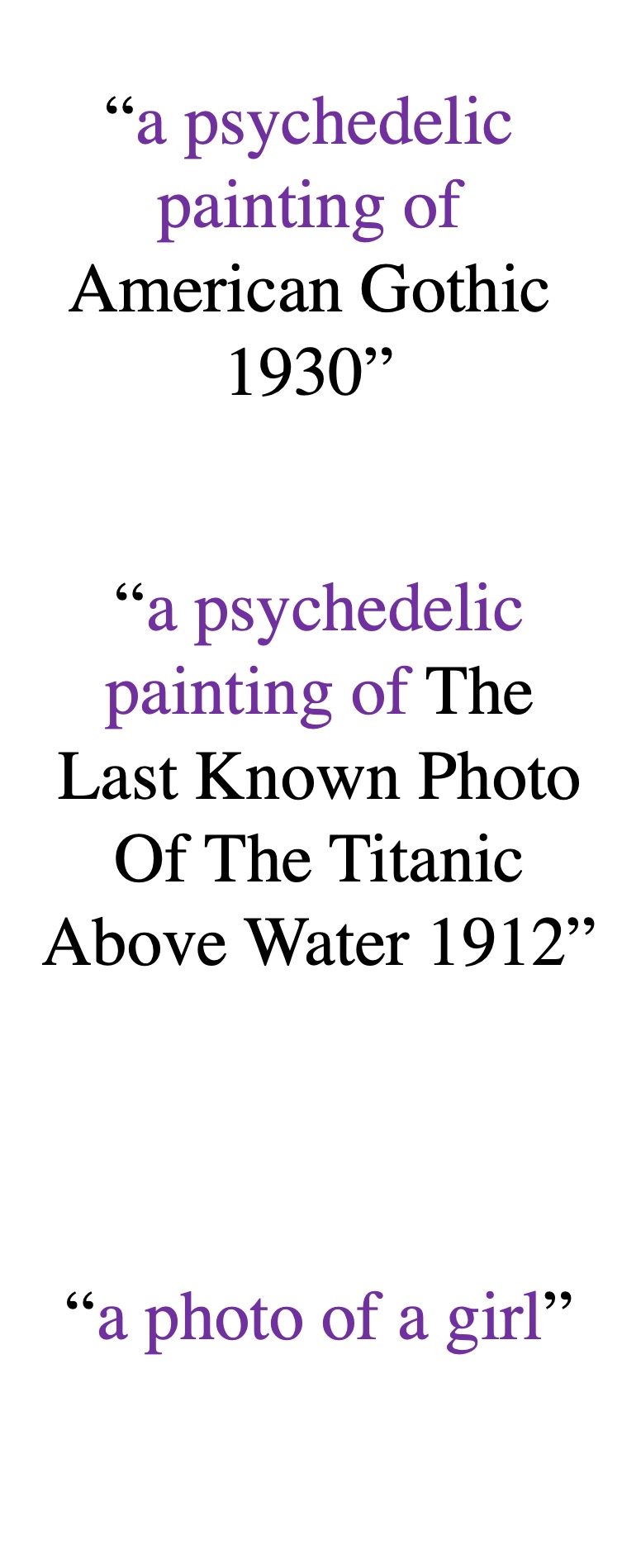} &
        \includegraphics[height=7.2cm]{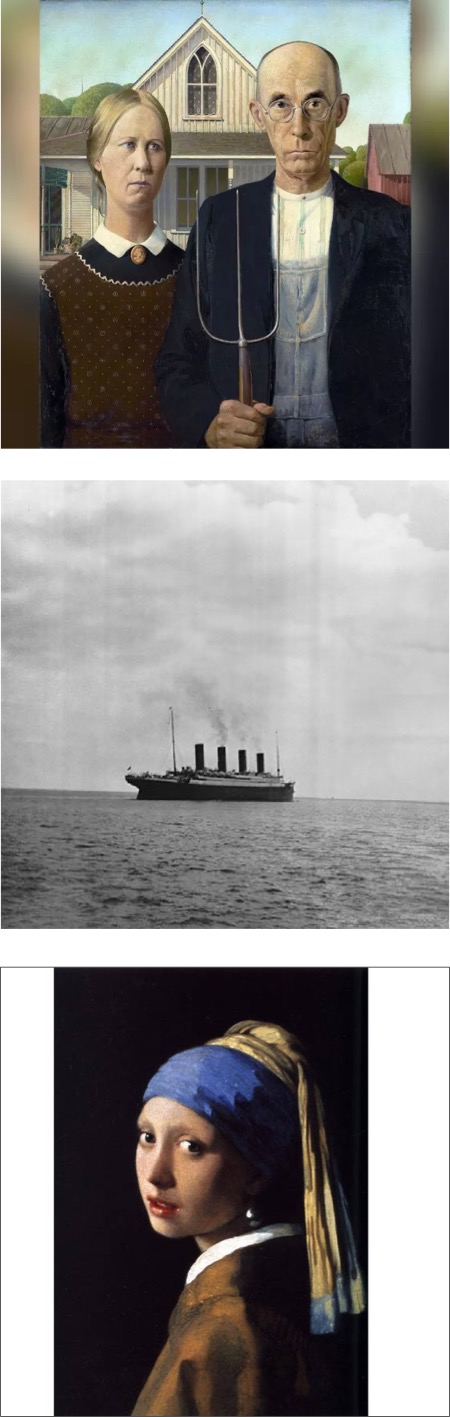} &
        \includegraphics[height=7.2cm]{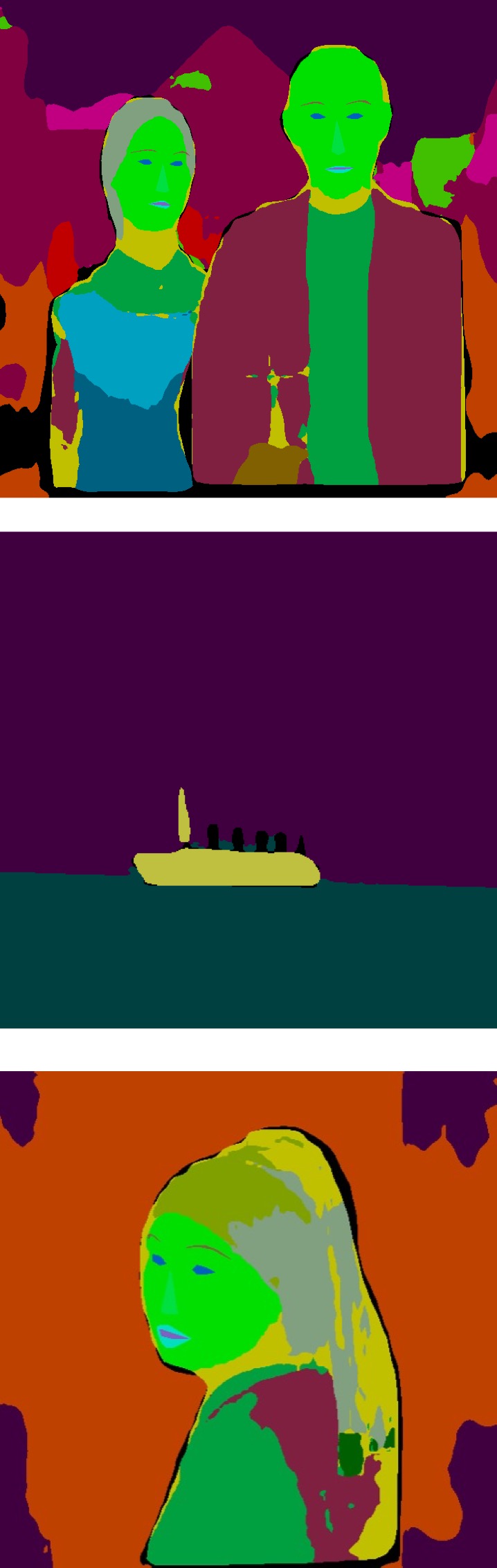} &
        \includegraphics[height=7.2cm]{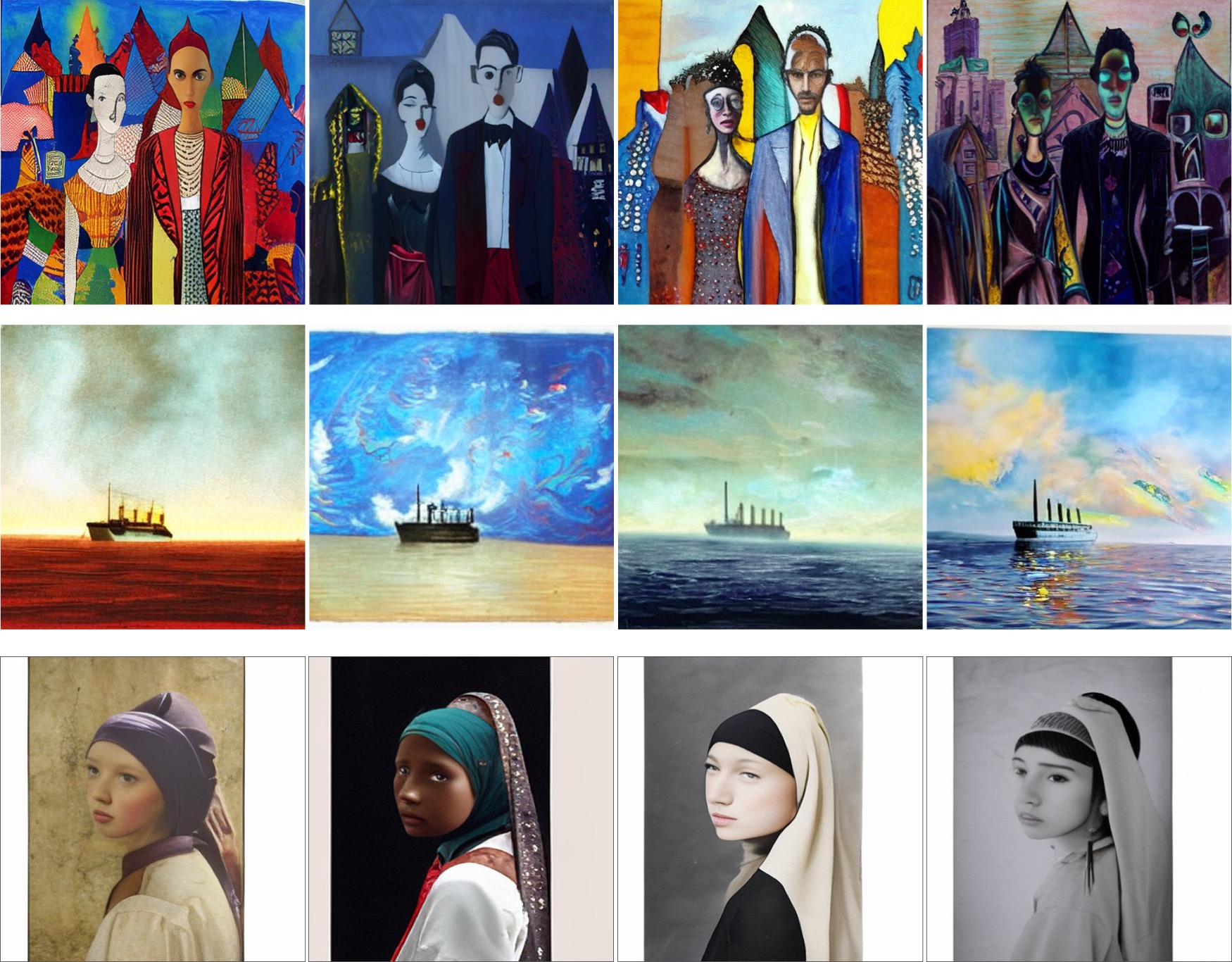} \\
        (a) & (b) & (c) & (d) \\
    \end{tabular}
    \caption{Generating new image interpretations through text editing and anchor scenes. For an input text (a) and image (b), we first extract the semantic segmentation (c), we can then re-generate new images (d) given the input segmentation and edited text. Purple denotes text added or replacing the original text.}
    \label{fig:edit_text}
    \vskip -0.1in
\end{figure*}

\subsection{Image-to-image generation} Generating images from segmentation maps or scenes can be viewed as a conditional image synthesis task~\cite{zhu2017unpaired,liu2017unsupervised,isola2017image,wang2018video,wang2018high,park2019semantic}. Specifically, this form of image synthesis permits more controllability over the desired output. CycleGAN~\cite{zhu2017unpaired} trained a mapping function from one domain to the other. UNIT~\cite{liu2017unsupervised} projected two different domains into a shared latent space and used a per-domain decoder to re-synthesize images in the desired domain. Both methods do not require supervision between domains. pix2pix~\cite{isola2017image} utilized conditional GANs together with a supervised reconstruction loss. pix2pixHD~\cite{wang2018high} improved the latter by increasing output image resolution thanks to improved network architecture. SPADE~\cite{park2019semantic} introduced a spatially-adaptive normalization layer which elevated information lost in normalization layers. ~\cite{gafni2021single} introduced face-refinement to SPADE through a pre-trained face-embedding network inspired by face-generation methods~\cite{gafni2019live}. Unlike the aforementioned, our work conditions jointly on text and segmentation, enabling bi-domain controllability.

\subsection{Text-to-image generation} Text-to-image generation~\cite{xu2018attngan,zhu2019dm,tao2020df,ye2021improving,zhang2021cross,ramesh2021zero,ding2021cogview,nichol2021glide,zhou2021lafite} focuses on generating images from standalone text descriptions. Preliminary text-to-image methods conditioned RNN-based DRAW~\cite{gregor2015draw} on text~\cite{mansimov2015generating}. Text-conditioned GANs provided additional improvement~\cite{reed2016generative}. AttnGAN~\cite{xu2018attngan} introduced an attention component, allowing the generator network to attend to relevant words in the text. DM-GAN~\cite{zhu2019dm} introduced a dynamic memory component, while DF-GAN~\cite{tao2020df} employed a fusion block, fusing text information into image features. Contrastive learning further improved the results of DM-GAN~\cite{ye2021improving}, while XMC-GAN~\cite{zhang2021cross} used contrastive learning to maximize the mutual information between image and text. 

DALL-E~\cite{ramesh2021zero} and CogView~\cite{ding2021cogview} trained an autoregressive transformer~\cite{vaswani2017attention} on text and image tokens, demonstrating convincing zero-shot capabilities on the MS-COCO dataset. GLIDE~\cite{nichol2021glide} used diffusion models conditioned on images. Inspired by the high-quality unconditional images generation model, GLIDE employed guided inference with and without a classifier network to generate high-fidelity images. LAFITE~\cite{zhou2021lafite} employed a pre-trained CLIP~\cite{radford2021learning} model to project text and images to the same latent space, training text-to-image models without text data. Similarly to DALL-E and CogView, we train an autoregressive transformer model on text and image tokens. Our main contributions are introducing additional controlling elements in the form of a scene, improve the tokenization process, and adapt classifier-free guidance to transformers.

\section{Method}
Our model generates an image given a text input and an optional scene layout (segmentation map). As demonstrated in our experiments, by conditioning over the scene layout, our method provides a new form of implicit controllability, improves structural consistency and quality, and adheres to human preference (as assessed by our human evaluation study).
In addition to our scene-based approach, we extended our aspiration of improving the general and perceived quality with a better representation of the token space. We introduce several modifications to the tokenization process, emphasizing awareness of aspects with increased importance in the human perspective, such as faces and salient objects. To refrain from post-generation filtering and further improve the generation quality and text alignment, we employ classifier-free guidance.

We follow next with a detailed overview of the proposed method, comprised of (i) scene representation and tokenization, (ii) attending human preference in the token space with explicit losses, (iii) the scene-based transformer, and (iv) transformer classifier-free guidance. Aspects commonly used prior to this method are not extensively detailed below, whereas specific settings for all elements can be found in the appendix.

\subsection{Scene representation and tokenization}
The scene is composed of a union of three complementary semantic segmentation groups - panoptic, human, and face.
By combining the three extracted semantic segmentation groups, the network learns to both generate the semantic layout and condition on it while generating the final image. The semantic layout provides additional global context in an implicit form that correlates with human preference, as the choice of categories within the scene groups, and the choice of the groups themselves are a prior to human preference and awareness. We consider this form of conditioning to be implicit, as the network may disregard any scene information, and generate the image conditioned solely on text. Our experiments indicate that both the text and scene firmly control the image.

In order to create the scene token space, we employ VQ-SEG: a modified VQ-VAE for semantic segmentation, building on the VQ-VAE suggested for semantic segmentation in~\cite{esser2021taming}. 
In our implementation the inputs and outputs of VQ-SEG are $m$ channels, representing the number of classes for all semantic segmentation groups $m=m_p+m_h+m_f+1$, where $m_p$, $m_h$, $m_f$ are the number of categories for the panoptic segmentation~\cite{wu2019detectron2}, human segmentation~\cite{li2020self}, and face segmentation extracted with~\cite{bulat2017far} respectively. The additional channel is a map of the edges separating the different classes and instances. The edge channel provides both separations for adjacent instances of the same class, and emphasis on scarce classes with high importance, as edges (perimeter) are less biased towards larger categories than pixels (area).

\subsection{Adhering to human emphasis in the token space}
We observe an inherent upper-bound on image quality when generating images with the transformer, stemming from the tokenization reconstruction method. In other words, quality limitations of the VQ image reconstruction method inherently transfer to quality limitations on images generated by the transformer. To that end, we introduce several modifications to both the segmentation and image reconstruction methods. These modifications are losses in the form of emphasis (specific region awareness) and perceptual knowledge (feature-matching over task-specific pre-trained networks). 

\subsection{Face-aware vector quantization} 
While using a scene as an additional form of conditioning provides an implicit prior for human preference, we institute explicit emphasis in the form of additional losses, explicitly targeted at specific image regions. 

We employ a feature-matching loss over the activations of a pre-trained face-embedding network, introducing ``awareness" of face regions and additional perceptual information, motivating high-quality face reconstruction. 

Before training the face-aware VQ (denoted as $\operatorname{VQ-IMG}$), faces are located using the semantic segmentation information extracted for VQ-SEG. The face locations are then used during the face-aware VQ training stage, running up to $k_f$ faces per image from the ground-truth and reconstructed images through the face-embedding network. The face loss can then be formulated as following:

\begin{equation}
    \mathcal{L}_\text{Face} = \sum_k\sum_l\alpha_f^l \|\operatorname{FE}^l(\hat{c}_{f}^{k})-\operatorname{FE}^{l}(c_{f}^{k})\|,
    \label{eq:vggface}
\end{equation}
where the index $l$ is used to denote the size of the spatial activation at specific layers of the face embedding network FE~\cite{vggface2}, while the summation runs over the last layers of each block of size $112\times112$, $56\times56$, $28\times28$, $7\times7$, $1\times1$ ($1\times 1$ being the size of the top most block), $\hat{c}_{f}^{k}$ and $c_{f}^{k}$ are respectively the reconstructed and ground-truth face crops $k$ out of $k_f$ faces in an image, $\alpha_f^l$ is a per-layer normalizing hyperparameter, and $\mathcal{L}_\text{Face}$ is the face loss added to the VQGAN losses defined by~\cite{esser2021taming}.

\subsection{Face emphasis in the scene space}
While training the VQ-SEG network, we observe a frequent reduction of the semantic segmentations representing the face parts (such as the eyes, nose, lips, eyebrows) in the reconstructed scene. This effect is not surprising due to the relatively small number of pixels that each face part accounts for in the scene space. A straightforward solution would be to employ a loss more suitable for class imbalance, such as focal loss~\cite{lin2017focal}. However, we do not aspire to increase the importance of classes that are both scarce, and of less importance, such as fruit or a tooth-brush. 
Instead, we (1) employ a weighted binary cross-entropy face loss over the segmentation face parts classes, emphasizing higher importance for face parts, and (2) include the face parts edges as part of the semantic segmentation edge map mentioned above.
The weighted binary cross-entropy loss can then be formulated as following:

\begin{equation}
    \mathcal{L}_\text{WBCE} = \alpha_{cat}\operatorname{BCE}(s,\hat{s}),
    \label{eq:wbce}
\end{equation}
where $s$ and $\hat{s}$ are the input and reconstructed segmentation maps respectively, $\alpha_{cat}$ is a per-category weight function, $\operatorname{BCE}$ is a binary cross-entropy loss, and $\mathcal{L}_\text{WBCE}$ is the weighted binary cross-entropy loss added to the conditional VQ-VAE losses defined by~\cite{esser2021taming}.

\subsection{Object-aware vector quantization}
We generalized and extend the face-aware VQ method to increase awareness and perceptual knowledge of objects defined as ``things" in the panoptic segmentation categories. Rather than a specialized face-embedding network, we employ a pre-trained VGG~\cite{simonyan2014very} network trained on ImageNet~\cite{krizhevsky2012imagenet}, and introduce a feature-matching loss representing the perceptual differences between the object crops of the reconstructed and ground-truth images. By running the feature-matching over image crops, we are able to increase the output image resolution from $256\times256$ by simply adding to VQ-IMG an additional down-sample and up-sample layer to the encoder and decoder respectively. Similarly to Eq.~\ref{eq:vggface}, the loss can be formulated as:

\begin{equation}
    \mathcal{L}_\text{Obj} = \sum_k\sum_l\alpha_o^l
    \|\operatorname{VGG}^l(\hat{c}_{o}^{k})-\operatorname{VGG}^{l}(c_{o}^{k})\|,
    \label{eq:vgg_obj}
\end{equation}
where $\hat{c}_{o}^{k}$ and $c_{o}^{k}$ are the reconstructed and input object crops respectively, $\operatorname{VGG}^l$ are the activations of the $l-th$ layer from the pre-trained $\operatorname{VGG}$ network, $\alpha_o^l$ is a per-layer normalizing hyperparameter, and $\mathcal{L}_\text{Obj}$ is the object-aware loss added to the VQ-IMG losses defined in Eq.~\ref{eq:vggface}.

\subsection{Scene-based transformer}
\begin{figure}[h]
    \centering
    \includegraphics[height=5.8cm]{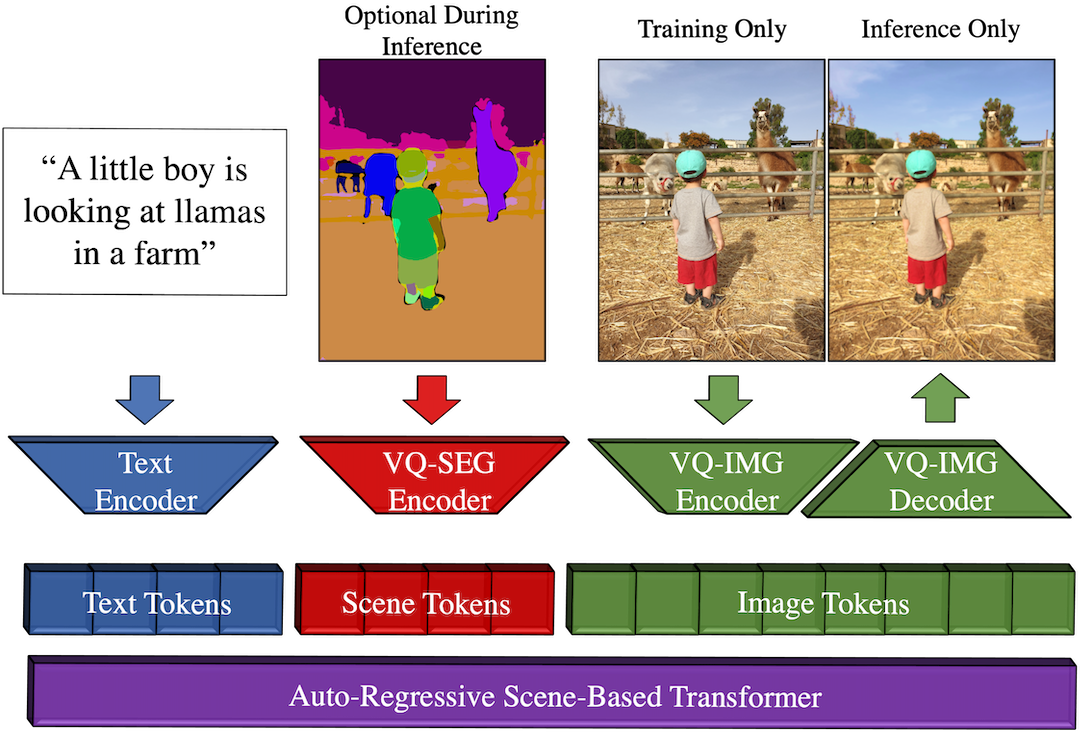}
    \caption{The scene-based method high-level architecture. Given an input text and optional scene layout, a corresponding image is generated. The transformer generates the relevant tokens, encoded and decoded by the corresponding networks.}
    \label{fig:arch_scene_v1}
\end{figure}

The method relies on an autoregressive transformer with three independent consecutive token spaces: text, scene, and image, as depicted in Fig~\ref{fig:arch_scene_v1}. The token sequence is comprised of $n_{x}$ text tokens encoded by a BPE~\cite{sennrich2015neural} encoder, followed by $n_{y}$ scene tokens encoded by VQ-SEG, and $n_{z}$ image tokens encoded or decoded by VQ-IMG. 

Prior to training the scene-based transformer, each encoded token sequence corresponding to a [text, scene, image] triplet is extracted using the corresponding encoder, producing a sequence that consists of:
\begin{gather*}
t_x, t_y, t_z = \operatorname{BPE}(i_x), \operatorname{VQ-SEG}(i_y), \operatorname{VQ-IMG}(i_z),  \\
t = [t_x, t_y, t_z],
\end{gather*}
where $i_x, i_y, i_z$ are the input text, scene and image respectively, $i_x \in \mathbb{N}^{d_x}$, $d_x$ is the length of the input text sequence, $i_y \in \mathbb{R}^{h_y \times w_y \times m}$, $i_z \in \mathbb{R}^{h_z \times w_z \times 3}$, $h_y,w_y,h_z,w_z$ are the height and width dimensions of the scene and image inputs respectively, $\operatorname{BPE}$ is the Byte Pair Encoding encoder, $t_x, t_y, t_z$ are the text, scene and image input tokens respectively, and $t$ is the complete token sequence.

\subsection{Transformer classifier-free guidance}
Inspired by the high-fidelity of unconditional image generation models, we employ classifier-free guidance~\cite{transformerCF,ho2021classifier,radford2021learning}. Classifier-free guidance is the process of guiding an unconditional sample in the direction of a conditional sample. To support unconditional sampling we fine-tune the transformer while randomly replacing the text prompt with padding tokens with a probability of $p_{CF}$. During inference, we generate two parallel token streams: a conditional token stream conditioned on text, and an unconditional token stream conditioned on an empty text stream initialized with padding tokens. For transformers, we apply classifier-free guidance on logit scores:
\begin{gather*}
    \text{logits}_{cond} = T(t_y,t_z | t_x), \\
    \text{logits}_{uncond} = T(t_y, t_z | \emptyset), \\
    \text{logits}_{cf} = \text{logits}_{uncond} + \alpha_c \cdot (\text{logits}_{cond} - \text{logits}_{uncond}),
    \label{eq:cf}
\end{gather*}
where $\emptyset$ is the empty text stream, $logits_{cond}$ are logit scores outputted by the conditioned token stream, $logits_{uncond}$ are logit scores outputted by the unconditioned token stream, $\alpha_c$ is the guidance scale, $logits_{cf}$ is the guided logit scores used to sample the next scene or image token, $T$ is an autoregressive transformer based the GPT-3~\cite{brown2020language} architecture. Note that since we use an autoregressive transformer, we use $logits_{cf}$ to sample once and feed the same token~(image or scene) to the conditional and unconditional stream.

\section{Experiments}
Our model achieves state-of-the-art results in human-based and numerical metric comparisons. Samples supporting the qualitative advantage are provided in Fig.~\ref{fig:prev_work}. Additionally, we demonstrate new creative capabilities possible with this method's new form of controllability. Finally, to better assess the effect of each contribution, an ablation study is provided. 

Experiments were performed with a 4 billion parameter transformer, generating a sequence of $256$ text tokens, $256$ scene tokens, and $1024$ image tokens, that are then decoded into an image with a resolution of $256\times256$ or $512\times512$ pixels (depending on the model of choice).

\subsection{Datasets} The scene-based transformer is trained on a union of CC12m~\cite{changpinyo2021conceptual}, CC~\cite{sharma2018conceptual}, and subsets of YFCC100m~\cite{thomee2016yfcc100m} and Redcaps~\cite{desai2021redcaps}, amounting to $35$m text-image pairs. MS-COCO~\cite{lin2014microsoft} is used unless otherwise specified. VQ-SEG and VQ-IMG are trained on CC12m, CC, and MS-COCO.

\subsection{Metrics} The goal of text-to-image generation is to generate high-quality and text-aligned images from a human perspective. Different metrics have been suggested to mimic the human perspective, where some are considered more reliable than others. We consider human evaluation the highest authority when evaluating image quality and text-alignment, and rely on FID~\cite{heusel2017gans} to increase evaluation confidence and handle cases where human evaluation is not applicable. We do not use IS~\cite{salimans2016improved} as it has been noted to be insufficient for model evaluation~\cite{barratt2018note}.

\subsection{Comparison with previous work}
\label{sec:prev_work}

The task of text-to-image generation does not contain absolute ground-truths, as a specific text description could apply to multiple images and vice versa. This constrains evaluation metrics to evaluate distributions of images, rather than specific images, thus we employ FID~\cite{heusel2017gans} as our secondary metric. 

\subsection{Baselines} We compare our results with several state-of-the-art methods using the FID metric and human evaluators (AMT) when possible. \textbf{DALL-E}~\cite{ramesh2021zero} provides strong zero-shot capabilities, similarly employing an autoregressive transformer with VQ-VAE tokenization. We train a re-implementation of DALL-E with $4$B parameters to enable human evaluation and fairly compare both methods employing an identical VQ method (VQGAN). \textbf{GLIDE}~\cite{nichol2021glide} demonstrates vastly improved results over DALL-E, adopting a diffusion-based~\cite{sohl2015deep} approach with classifier-free guidance~\cite{ho2021classifier}. We additionally provide an FID comparison with \textbf{CogView}~\cite{ding2021cogview}, \textbf{LAFITE}~\cite{zhou2021lafite}, \textbf{XMC-GAN}~\cite{zhang2021cross}, \textbf{DM-GAN(+CL)}~\cite{ye2021improving}, \textbf{DF-GAN}~\cite{tao2020df}, \textbf{DM-GAN}~\cite{zhu2019dm}, \textbf{DF-GAN}~\cite{tao2020df} and, \textbf{AttnGAN}~\cite{xu2018attngan}.

\subsection{Human evaluation results} Human evaluation with previous methods is provided in Tab.~\ref{table:prev_work}. In each instance, human evaluators are required to choose between two images generated by the two models being compared. The two models are compared in three aspects: (i) image quality, (ii) photorealism (which image appears more real), and (iii) text alignment (which image best matches the text). Each question is surveyed using $500$ image pairs, where $5$ different evaluators answer each question, amounting to $2500$ instances per question for a given comparison.
We compare our $256\times256$ model with our re-implementation of DALL-E~\cite{ramesh2021zero} and CogView's~\cite{ding2021cogview} $256\times256$ model. CogView's $512\times512$ model is compared with our corresponding model. Results are presented as a percentage of majority votes in favor of our method when comparing between a certain model and ours. Compared with the three methods, ours achieves significantly higher favorability in all aspects.

\subsection{FID comparison} FID is calculated over a subset of $30k$ images generated from the MS-COCO validation set text prompts with no re-ranking, and provided in Tab.~\ref{table:prev_work}. The evaluated models are divided into two groups: trained with and without (denoted as filtered) the MS-COCO training set. In both scenarios our model achieves the lowest FID. In addition, we provide a loose practical lower-bound (denoted as ground-truth), calculated between the training and validation
subsets of MS-COCO. As FID results are approaching small numbers, it is interesting to get an idea of a possible practical lower-bound.

\setlength{\tabcolsep}{1.5pt}
\begin{table}
\begin{center}
\label{table:prev_work}
\begin{tabular}{l|cc|ccc}
\hline\noalign{\smallskip}
Model & FID$\downarrow$ & FID$\downarrow$ & Image & Photo- & Text \\
& & (filt.) & quality & realism & alignment  \\
\noalign{\smallskip}
\hline
\noalign{\smallskip}
AttnGAN~\cite{xu2018attngan}  & $35.49$ & - & - & - & - \\
DM-GAN~\cite{zhu2019dm} & $32.64$ & - & - & - & - \\
DF-GAN~\cite{tao2020df} & $21.42$ & - & - & - & - \\
DM-GAN+CL~\cite{ye2021improving} & $20.79$ & - & - & - & -\\
XMC-GAN~\cite{zhang2021cross} & $9.33$ & - & - & - & -\\
DALL-E~\cite{ramesh2021zero} & - & $34.60$ & $81.8\%$ & $81.0\%$ & $65.9\%$ \\
CogView$_{256}$~\cite{ding2021cogview} & - & $32.20$  & $92.2\%$ & $94.2\%$ & $92.2\%$\\
CogView$_{512}$~\cite{ding2021cogview} & - & $36.53$ & $91.1\%$  & $88.2\%$ & $87.8\%$ \\
LAFITE~\cite{zhou2021lafite} & $8.12$ & $26.94$ & - & - & - \\
GLIDE~\cite{nichol2021glide} & - & $12.24$ & - &  - & - \\
\bf{Ours}$_{256}$ & {$\bf{7.55}$} & $\bf{11.84}$ \\
\hline
Ground-truth & $2.47$ & - & - & - & - \\
\hline
\end{tabular}
\vspace{0.2cm}
\caption{Comparison with previous work (FID and human preference). FID is calculated over a subset of $30k$ images generated from the MS-COCO validation set text prompts. When possible, we include models trained with and without (filtered) the MS-COCO training set. In both scenarios our model achieves state of the art results, correlating with visual samples and human evaluation. We add a loose practical lower-bound (denoted as ground-truth), calculated between the training and validation subsets of MS-COCO. Human evaluation is shown as a percentage of majority votes in favor of our method when comparing between a certain model and ours.}
\end{center}
\end{table}
\setlength{\tabcolsep}{1.4pt}

\subsection{Generating out of distribution}
Methods that rely on text inputs only are more confined to generate within the training distribution, as demonstrated by~\cite{nichol2021glide}. Unusual objects and scenarios can be challenging to generate, as certain objects are strongly correlated with specific structures, such as cats with four legs, or cars with round wheels. The same is true for scenarios. ``A mouse hunting a lion" is most likely not a scenario easily found within the dataset. By conditioning on scenes in the form of simple sketches, we are able to attend to these uncommon objects and scenarios, as demonstrated in Fig.~\ref{fig:attending}, despite the fact that some objects do not exist as categories in our scene (mouse, lion). We solve the category gap by using categories that may be close in certain aspects (elephant instead of mouse, cat instead of lion). In practice, for non-existent categories, several categories could be used instead.

\subsection{Scene controllability}
Samples are provided in Fig.~\ref{fig:teaser},~\ref{fig:attending},~\ref{fig:edit_scene},~\ref{fig:edit_text} and in the appendix with both our $256\times256$ and $512\times512$ models. In addition to generating high fidelity images from text only, we demonstrate the applicability of scene-wise image control and maintaining consistency between generations.

\subsection{Scene editing and anchoring} 
Rather than editing certain regions of images as demonstrated by~\cite{ramesh2021zero}, we introduce new capabilities of generating images from existing or edited scenes. In Fig.~\ref{fig:edit_scene}, two scenarios are considered. In both scenarios the semantic segmentation is extracted from an input image, and used to re-generate an image conditioned on the input text. In the top row, the scene is edited, replacing the `sky' and `tree' categories with `sea', and the `grass' category with `sand', resulting in a generated image adhering to the new scene. A simple sketch of a giant dog is added to the scene in the bottom row, resulting in a generated image corresponding to the new scene without any change in text.

Fig.~\ref{fig:edit_text} demonstrates the ability to generate new interpretations of existing images and scenes. After extracting the semantic segmentation from a given image, we re-generate the image conditioned on the input scene and edited text.

\subsection{Storytelling through controllability} 
To demonstrate the applicability of harnessing scene control for story illustrations, we wrote a children story, and illustrated it using our method. The main advantages of using simple sketches as additional inputs in this case, are (i) that authors can translate their ideas into paintings or realistic images, while being less susceptible to the ``randomness" of text-to-image generation, and (ii) improved consistency between generation. We provide a \href{https://youtu.be/QLTyqoJJKTo}{short video} of the story and process.

\subsection{Ablation study}
An ablation study of human preference and FID is provided in Tab.~\ref{table:abl_study} to assess the effectiveness of our different contributions. Settings in both studies are similar to the comparison made with previous work (Sec.~\ref{sec:prev_work}). Each row corresponds to a model trained with the additional element, compared with the model without that specific addition for human preference. We note that while the lowest FID is attained by the $256\times256$ model, human preference favors the $512\times512$ model with object-aware training, particularly in quality. Furthermore, we re-examine the FID of the best model, where the scene is given as an additional input, to gain a better notion of the gap from the lower-bound (Tab.~\ref{table:prev_work}).

\setlength{\tabcolsep}{4pt}
\begin{table}
\begin{center}
\label{table:abl_study}
\begin{tabular}{lc|ccc}
\hline\noalign{\smallskip}
Model & FID$\downarrow$ & Image & Photo- & Text \\
& & quality & realism & alignment  \\
\noalign{\smallskip}
\hline
\noalign{\smallskip}
Base & $18.01$ & - & - & - \\
$+$Scene tokens & $19.16$ & $57.3\%$  & $65.3\%$ & $58.3\%$ \\
$+$Face-aware & $14.45$ & $63.6\%$  & $59.8\%$ & $57.4\%$ \\
$+$\textbf{CF} & $\textbf{7.55}$ & $76.8\%$  & $66.8\%$ & $66.8\%$ \\
$+$\textbf{Obj-aware}$_{512}$ & $8.70$ & $62.0\%$  & $53.5\%$ & $52.2\%$ \\ 
\hline
$+$CF with scene input & $4.69$ & - & - & - \\
\hline
\end{tabular}
\vspace{0.2cm}
\caption{Ablation study (FID and human preference). FID is calculated over a subset of $30k$ images generated from the MS-COCO validation set text prompts. Human evaluation is shown as a percentage of majority votes in favor of the added element compared to the previous model.}

\end{center}
\end{table}
\setlength{\tabcolsep}{1.4pt}

\section{Conclusion}
The text-to-image domain has witnessed a plethora of novel methods aimed at improving the general quality and adherence to text of generated images. While some methods propose image editing techniques, progress is not often directed towards enabling new forms of human creativity and experiences. We attempt to progress text-to-image generation towards a more interactive experience, where people can perceive more control over the generated outputs, thus enable real-world applications such as storytelling. In addition to improving the general image quality, we focus on improving key image aspects we deem significant in human perception, such as faces and salient objects, resulting in higher favorability of our method in human evaluations and objective metrics.

{\small
\bibliographystyle{ieee_fullname}
\bibliography{egbib}
}
\clearpage
\appendix

\section{Additional implementation details}

\subsection{VQ-SEG}
VQ-SEG is trained for $600k$ iterations, with a batch size of $48$, dictionary size of $1024$. The number of segmentation categories per-group are $m_p=133$ for the panoptic segmentation, $m_h=20$ for the human parsing, and $m_f=5$ for the face parsing. The per-category weight function follows the notation:

\begin{equation}
    \alpha_{cat}=
    \begin{cases}
      20, & \text{if}\ \text{cat} \in [154,...,158] \\
      1, & \text{otherwise},
    \end{cases}
  \end{equation}
where $\text{cat} \in [154,...,158]$ are the face-parts categories eyebrows, eyes, nose, outer-mouth, and inner-mouth.

\subsection{VQ-IMG}
$\operatorname{VQ-IMG_{256}}$ and $\operatorname{VQ-IMG_{512}}$ are trained for $800k$ and $940k$ iterations respectively, with a batch size of $192$ and $128$, a channel multiplier of $[1,1,2,4]$ and $[1,1,2,4,4]$, while both are trained with a dictionary size of $8192$. 

The per-layer normalizing hyperparameter for the face-aware loss is $\alpha_f^l=[\alpha_{f1}, \alpha_{f2}\times0.01, \alpha_{f2}\times0.1, \alpha_{f2}\times0.2, \alpha_{f2}\times0.02]$ corresponding to the last layer of each block of size $1\times1, 7\times7, 28\times28, 56\times56, 128\times128$, where $\alpha_{f1}=0.1$ and $\alpha_{f2}=0.25$. We experimented with two settings, the first where $\alpha_{f1}=\alpha_{f2}=1.0$, and the second, which was used to train the final models, where $\alpha_{f1}=0.1,\alpha_{f2}=0.25$. The remaining face-loss values were taken from the work of~\cite{gafni2019live}.
The per-layer normalizing hyperparameter for the object-aware loss, $\alpha_o^l$ were taken from the work of~\cite{esser2021taming}, based on LPIPS~\cite{zhang2018unreasonable}.

\subsection{Scene-based transformer}
The $512\times512$ and $256\times256$ models both share all implementation details, excluding the VQ-IMG used for token encoding and decoding, and the object-aware loss that was applied to the $512\times512$ model only. Both transformers share the architecture of $48$ layers, $48$ attention heads, and an embedding dimension of $2560$. The models were trained for a total of $170k$ iterations, with a batch size of $1024$, Adam~\cite{kingma2014adam} optimizer, with a starting learning-rate of $4.5\times10^{-4}$ for the first $40k$ iterations, transitioning to $1.5\times10^{-4}$ for the remainder, $\beta_1=0.9$,$\beta_2=0.96$, weight-decay of $4.5\times10^{-4}$, and a loss ratio of $7/1$ between the image and text tokens.
For classifier-free guidance, we fine-tune the transformer, while replacing the text tokens with padding tokens in the last $30k$ iterations, with a probability of $p_{CF}=0.2$. At inference-time we set the guidance scale to $\alpha_c=5$, though we found that $\alpha_c=3$ works as well.

At each inference step, the next token is sampled by (i) selecting half the logits with the highest probabilities, (ii) applying a softmax operation over the selected logits, and (iii) sampling a single logit from a multinomial probability distribution.

\section{Additional samples}
\begin{figure*}[t]
    \centering
    \includegraphics[height=22cm]{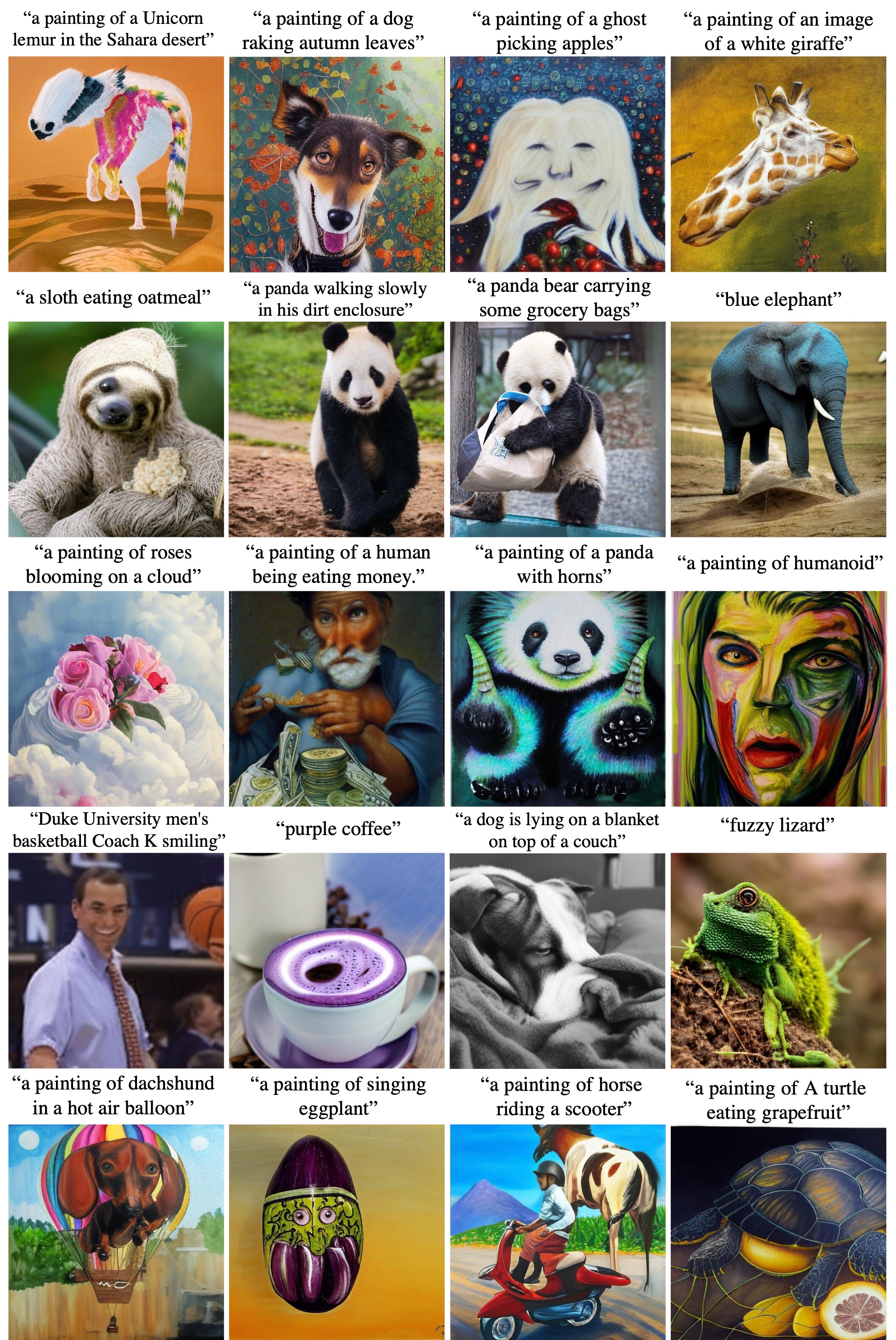}
    \caption{Additional samples generated from challenging text inputs.}
    \label{fig:more_samples_1}
\end{figure*}

\begin{figure*}[t]
    \centering
    \includegraphics[height=22cm]{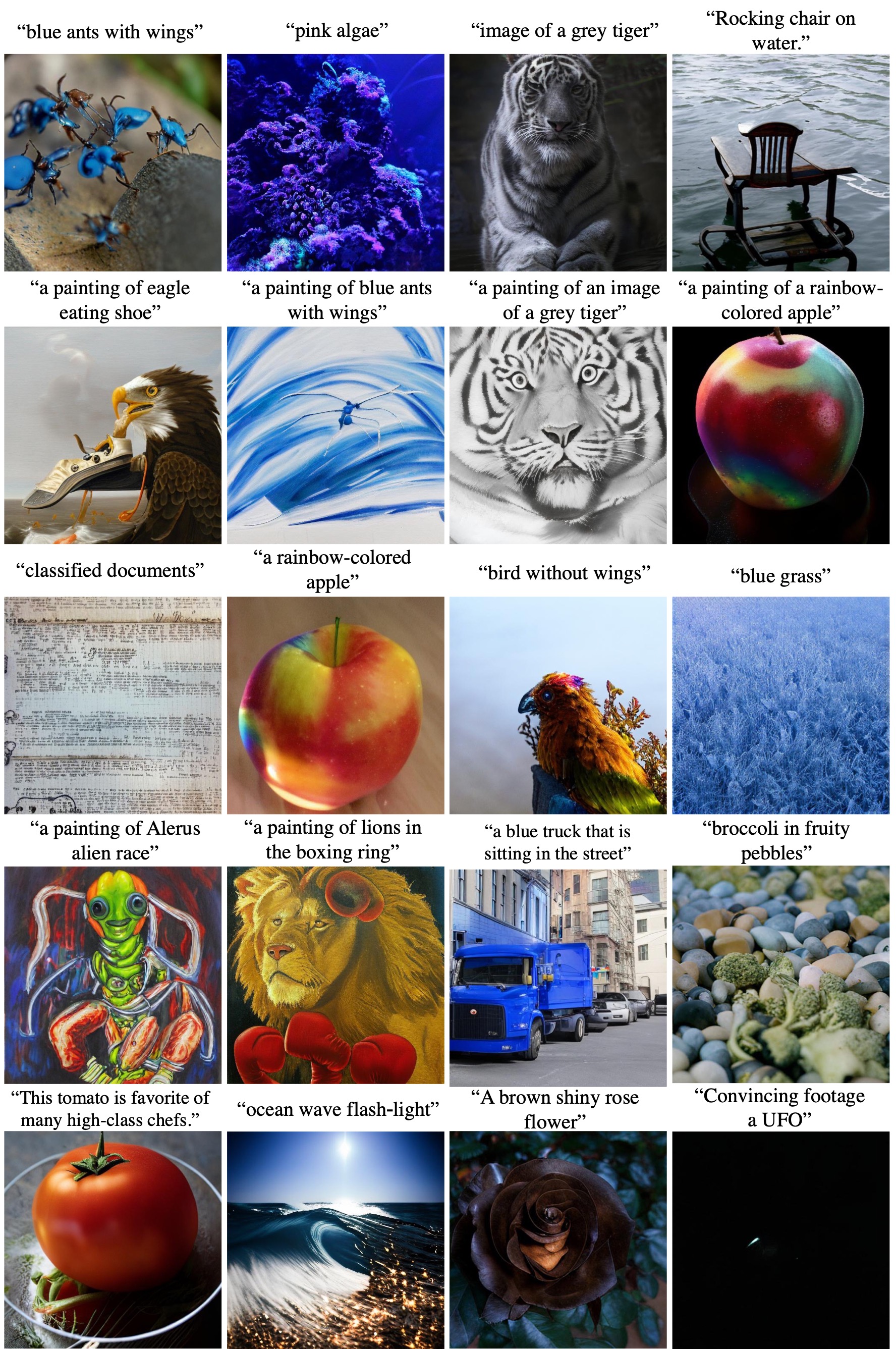}
    \caption{Additional samples generated from challenging text inputs.}
    \label{fig:more_samples_2}
\end{figure*}

\begin{figure*}[t]
    \centering
    \begin{tabular}{cc}
    \includegraphics[height=10cm]{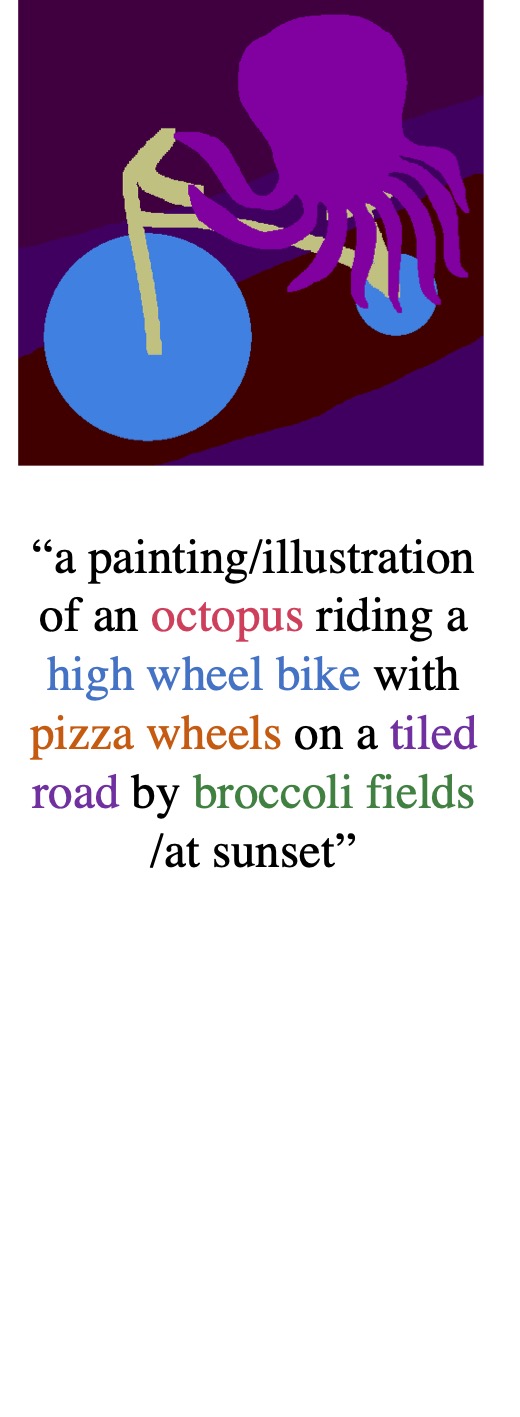} & \includegraphics[height=10cm]{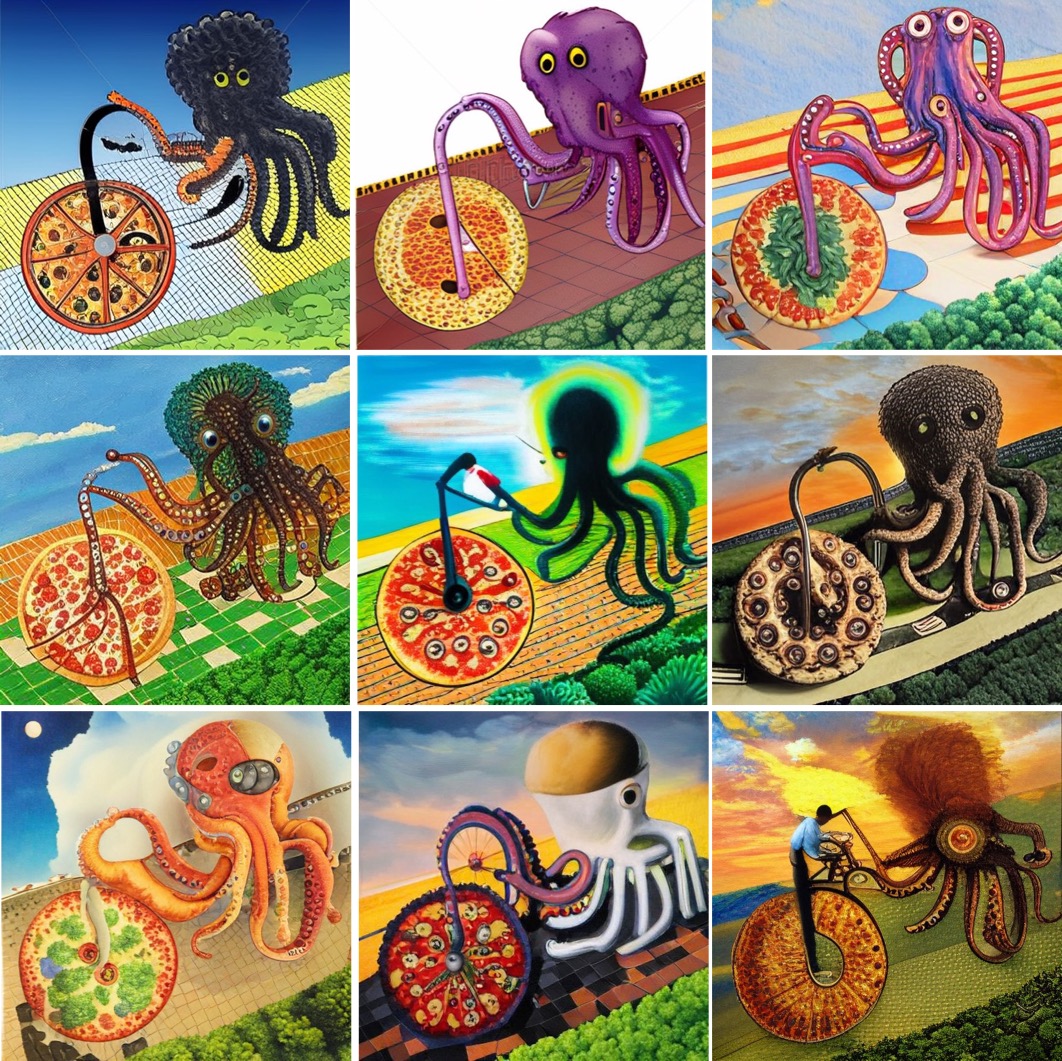} \\
        (a) & (b) 
    \end{tabular}
    \caption{Additional samples generated (b) from text and segmentation inputs (a).}
    \label{fig:more_seg_samples_1}
\end{figure*}

\begin{figure*}[t]
    \centering
    \begin{tabular}{cc}
    \includegraphics[height=10cm]{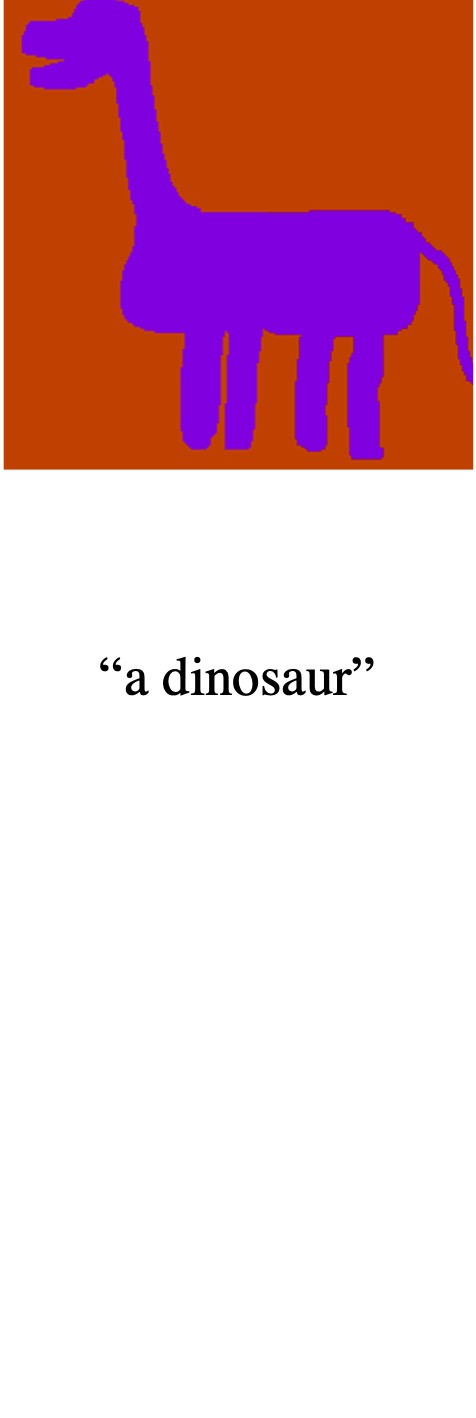} & \includegraphics[height=10cm]{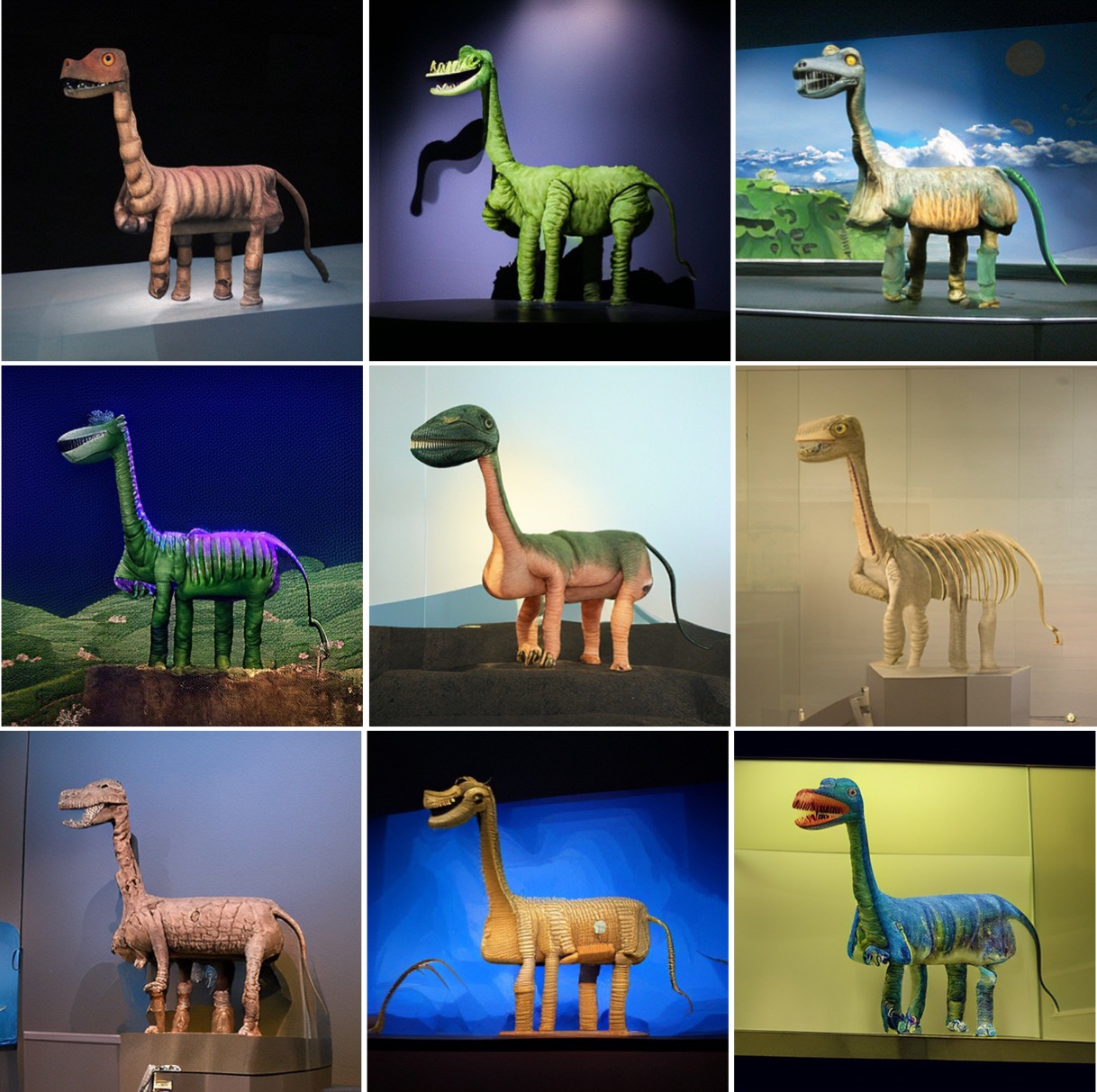} \\
        (a) & (b) 
    \end{tabular}
    \caption{Additional samples generated (b) from text and segmentation inputs (a).}
    \label{fig:more_seg_samples_2}
\end{figure*}

\begin{figure*}[t]
    \centering
    \begin{tabular}{cc}
    \includegraphics[height=10cm]{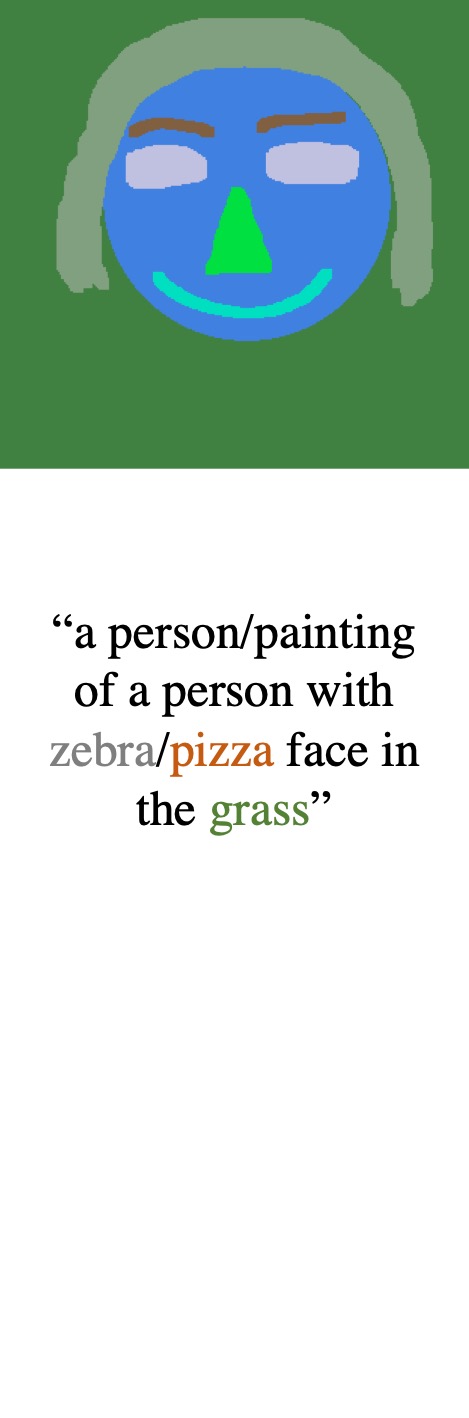} & \includegraphics[height=10cm]{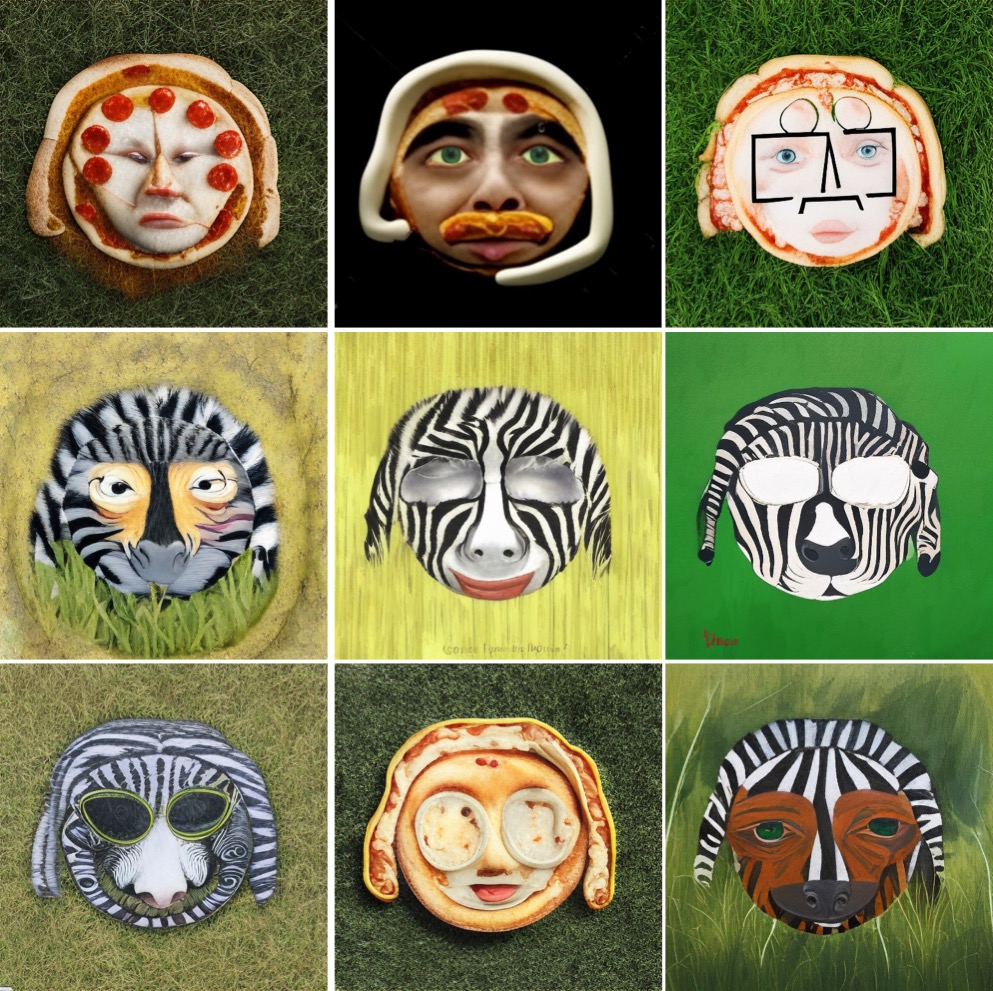} \\
        (a) & (b) 
    \end{tabular}
    \caption{Additional samples generated (b) from text and segmentation inputs (a).}
    \label{fig:more_seg_samples_3}
\end{figure*}

\begin{figure*}[t]
    \centering
    \begin{tabular}{cc}
    \includegraphics[height=10cm]{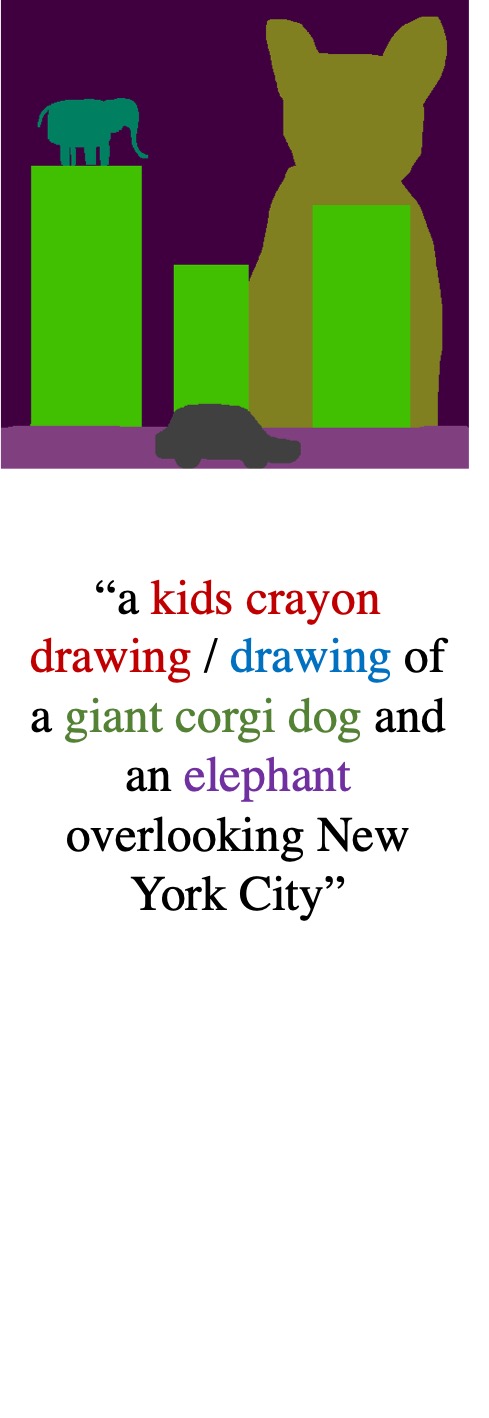} & \includegraphics[height=10cm]{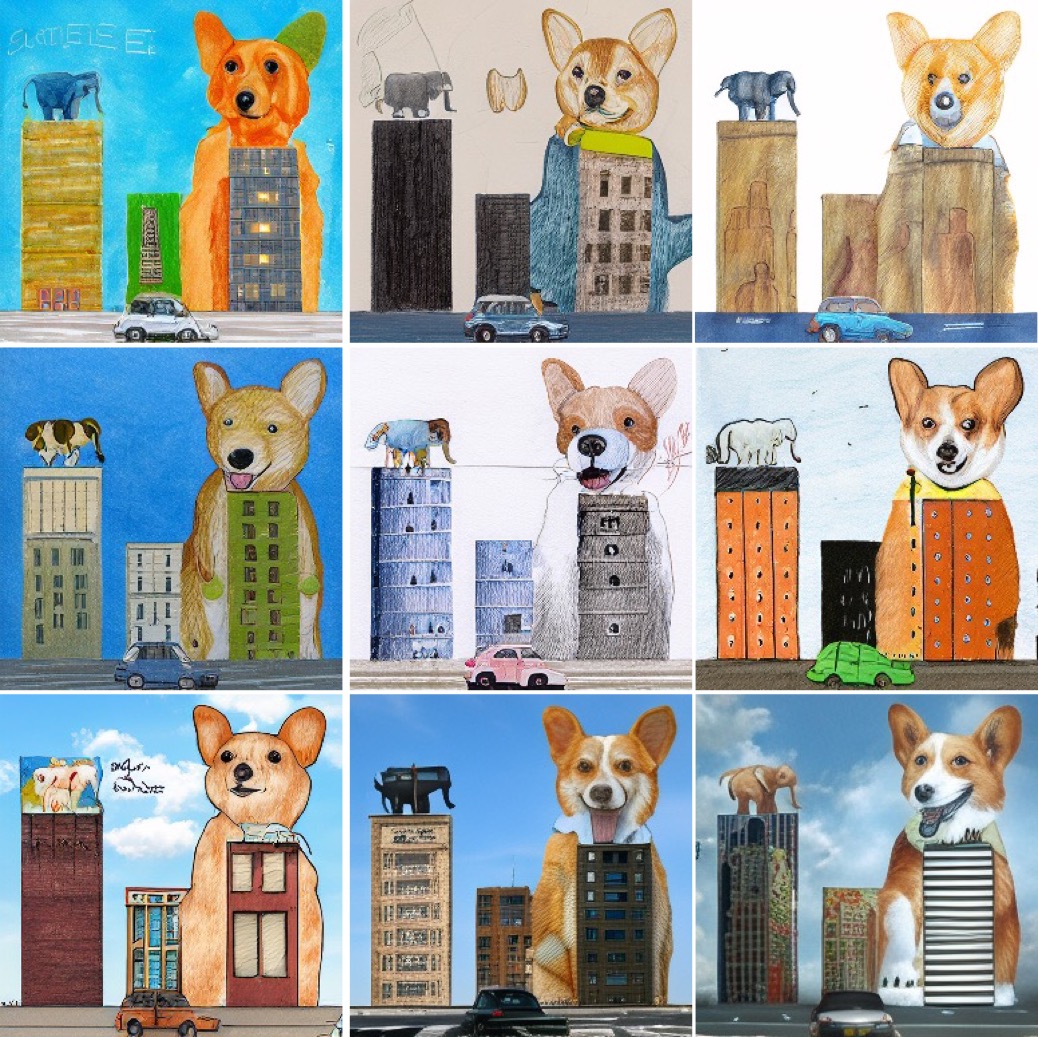} \\
        (a) & (b) 
    \end{tabular}
    \caption{Additional samples generated (b) from text and segmentation inputs (a).}
    \label{fig:more_seg_samples_4}
\end{figure*}

Additional samples generated from challenging text inputs are provided in Figs.~\ref{fig:more_samples_1}-\ref{fig:more_samples_2}, while samples generated from text and scene inputs are provided in Figs.~\ref{fig:more_seg_samples_1}-\ref{fig:more_seg_samples_4}. The different text colors emphasize the large number of different objects/scenarios being attended. As there are no `octopus' or `dinosaur' categories, we use instead the `cat' and `giraffe' categories respectively. We did not attempt to use other classes in this case. However, we found that generally there are no ``one-to-one'' mappings between absent and existing categories, hence several categories may work for an absent category.

\end{document}